\pdfoutput=1
\documentclass[11pt]{article}

\usepackage[final]{acl}

\usepackage{booktabs}
\usepackage{times}
\usepackage{latexsym}  

\usepackage[utf8]{inputenc}
\usepackage[T1]{fontenc}
\usepackage{microtype}

\usepackage{inconsolata}

\usepackage{amsmath,amsfonts}
\DeclareMathOperator*{\argmax}{arg\,max}

\usepackage{array}
\usepackage[caption=false,font=normalsize,labelfont=sf,textfont=sf]{subfig}
\usepackage{textcomp}
\usepackage{stfloats}
\usepackage{url}
\usepackage{verbatim}
\usepackage{graphicx}
\usepackage{tabularx}
\usepackage{multirow}
\usepackage{hyperref}
\usepackage{afterpage}
\usepackage{float}
\usepackage{enumitem}    % For custom bullet styles
\usepackage{amssymb}     % For symbols like \checkmark
\usepackage{pifont}      % For more bullet symbols (like \ding)
\usepackage{newunicodechar}

\newunicodechar{á}{\'a}
\newunicodechar{é}{\'e}
\newunicodechar{í}{\'i}
\newunicodechar{ó}{\'o}
\newunicodechar{ú}{\'u}

\newunicodechar{ẹ}{\d{e}}
\newunicodechar{ọ}{\d{o}}
\newunicodechar{ị}{\d{i}}
\newunicodechar{ụ}{\d{u}}

\newunicodechar{ñ}{\~n}
\newunicodechar{ü}{\"u}

\DeclareUnicodeCharacter{0300}{\`{}}  % grave accent
\DeclareUnicodeCharacter{0301}{\'{}}  % acute accent

\usepackage{algorithm}
\usepackage{algpseudocode}

\usepackage[most]{tcolorbox}
\usepackage[dvipsnames]{xcolor}
\definecolor{grpo}{HTML}{FAD4D4}   % very light red/pink tone
\definecolor{gfpo}{HTML}{BEE4D0}   % cool green
\definecolor{comparison}{HTML}{B3D9FF} % light blue

\definecolor{grayrowcolor}{RGB}{220,220,220} % Light gray color

\definecolor{commentcolour}{rgb}{0.3,0.7,0.2}
\definecolor{lightblue}{RGB}{245, 250, 250} % lighter & softer
\definecolor{blue}{RGB}{77, 174, 172} % softer teal

\definecolor{darkpink}{RGB}{255, 105, 180} 

\newtcolorbox{findings}{
    enhanced,
    breakable,
    colback=lightblue,
    colframe=blue,
    boxrule=1.5pt,
    arc=0.25em,
    left=1em,
    right=1em,
    top=1em,
    bottom=0.75em,
    before=\vspace{1em},
    overlay unbroken and first={
        \node[
            fill=blue,
            text=white,
            font=\bfseries,
            anchor=west,
            inner xsep=0.75em,
            inner ysep=0.5em,
            rounded corners=0.25em
        ] 
        at ([xshift=0.75em]frame.north west) {Finding};
    }
}

\newtcolorbox[auto counter]{prompt}[3][]{
  enhanced,
  breakable,
  colback=#2!10,
  colframe=#2!50!black,
  title=\textbf{#3},
  fontupper=\normalsize\fontfamily{cmss}\selectfont,
  #1
}

\newif\ifdraft
\drafttrue   % set to \draftfalse to hide comments

\ifdraft
  \newcommand{\prk}[1]{\textcolor{orange}{[\textbf{Prashant:} #1]}}
\else
  \newcommand{\prk}[1]{} % comments vanish
\fi

\definecolor{highlightblue}{RGB}{0,90,200}

\definecolor{correction}{RGB}{0,90,200}
\definecolor{previous}{RGB}{255,140,0}
 
\title{Neither Here Nor There: Cross-Lingual Representation Dynamics of Code-Mixed Text --- A Hindi–English Case Study}

\author{Debajyoti Mazumder$^1$, Divyansh Pathak$^1$, Prashant Kodali$^2$, Jasabanta Patro$^1$ \\ 
$^1$Indian Institute of Science Education and Research, Bhopal, India \hspace{0.3cm} $^2$Microsoft Research India \\
\texttt{debajyoti22@iiserb.ac.in, divyansh22@iiserb.ac.in,} \\ \texttt{kodali.prashant@gmail.com, jpatro@iiserb.ac.in}
}

\begin{document}
\maketitle
\begin{abstract}

Multilingual encoder-based language models are widely used for code-mixed analysis, yet their internal representations of code-mixed inputs—and their relationship to the constituent languages—remain poorly understood. Using Hindi–English as a case study, we construct a unified trilingual corpus of parallel English, Hindi (Devanagari), and Romanized code-mixed sentences. We then probe cross-lingual representation alignment in standard multilingual encoders and their code-mix-adapted variants using CKA, token-level saliency, and entropy-based uncertainty analysis. We find that while standard models align English and Hindi well, code-mixed inputs remain loosely connected to either language — and that continued pre-training on code-mixed data improves English–code-mixed alignment at the cost of English–Hindi alignment. Interpretability analyses further reveal a clear asymmetry: models process code-mixed text through an English-dominant semantic subspace, while native-script Hindi provides complementary signals that reduce representational uncertainty. Motivated by these findings, we introduce a trilingual post-training alignment objective that brings code-mixed representations closer to both constituent languages simultaneously, yielding more balanced cross-lingual alignment and downstream gains on sentiment analysis and hate speech detection — showing that grounding code-mixed representations in their constituent languages meaningfully helps cross-lingual understanding. Code is available at: \url{https://github.com/debajyotimaz/tri_align_EMNLP_2026}.

\end{abstract}
\section{Introduction}

\begin{figure}
    \centering
    \includegraphics[
        width=0.9\columnwidth,
        keepaspectratio
    ]{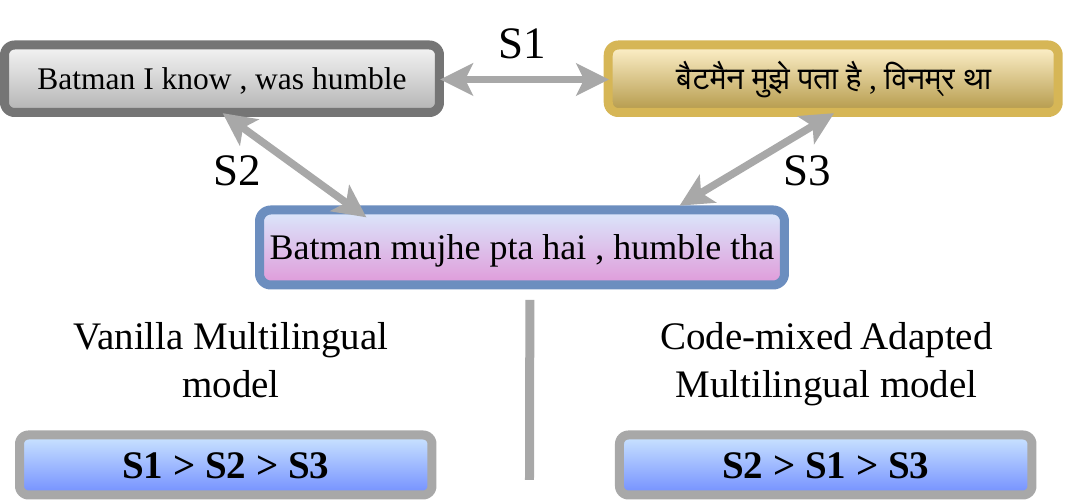}
    \caption{
        Vanilla multilingual model versus code-mixed adapted multilingual model. Here, S represents similarity.
    }
    \label{fig:idea}
\end{figure}
\begingroup
\renewcommand\thefootnote{}
\footnotetext{CM denotes code-mixed text. The terms \textit{participating} and \textit{constituent} languages are used interchangeably.}
\addtocounter{footnote}{-1}
\endgroup
Code-mixing is a linguistic phenomenon in multilingual societies where speakers combine words or phrases from multiple languages within a single sentence \cite{gumperz1982discourse}. Despite its prevalence, code-mixed text poses challenges for multilingual language models (MLMs), which are typically pretrained on predominantly monolingual corpora \cite{devlin-etal-2019-bert, NEURIPS2019_c04c19c2, conneau-etal-2020-unsupervised, zhang-etal-2023-multilingual}. A widely studied example is Hindi–English code-mixing, often written in Roman script and commonly referred to as Hinglish.

To support multilingual understanding, encoder-based MLMs such as mBERT and XLM-R learn shared representation spaces in which semantically equivalent sentences across languages tend to align \cite{10.1145/3764112}. Such cross-lingual alignment is crucial for multilingual NLP because it enables knowledge/capabilities learned in one language to transfer to others in downstream tasks such as sentiment analysis and hate speech detection \cite{de-varda-marelli-2024-emergence, hong-etal-2025-cross}. Code-mixed text provides a natural testbed for studying cross-lingual alignment because lexical and semantic cues from multiple constituent languages appear within a single utterance, requiring models to integrate signals from different languages into a coherent representation. Similar to how English can serve as a pivot language in multilingual models \cite{kargaran-etal-2025-mexa}, linguistically, code-mixed text rests on the \textit{twin foundations} of its constituent monolingual languages. However, most studies on cross-lingual alignment examine representations of monolingual sentences across languages, leaving it unclear how multilingual encoders align code-mixed inputs with their constituent monolingual languages \cite{hammerl-etal-2024-understanding}. As a result, the relationship between code-mixed representations and those of their corresponding monolingual counterparts remains underexplored. This lack of understanding limits our ability to design effective multilingual models that properly handle code-mixed inputs.

In this study, we examine the representational dynamics of multilingual encoders when processing code-mixed text relative to their monolingual counterparts. To enable this analysis, we construct a \textit{trilingual corpus} of parallel English, Hindi, and Hindi–English Code-Mixed (CM) sentences from existing resources. Our analysis examines how constituent language signals influence MLM understanding of code-mixed inputs using interpretability tools such as Centered Kernel Alignment (CKA), token-level saliency, and entropy-based uncertainty reduction.

Our observations reveal several key findings. First, while multilingual models align monolingual English and Hindi representations well, code-mixed representations remain weakly anchored to both languages. Second, continued pretraining on code-mixed data improves English–CM alignment but degrades English–Hindi alignment, revealing a trade-off between code-mixed adaptation and monolingual alignment. Third, interpretability analyses show that code-mixed representations are largely explained by the English representation space, while native-script Hindi contributes complementary signals that reduce representational uncertainty.

Building on these insights, we propose grounding code-mixed
representations in both constituent languages through trilingual
alignment across EN--HI, EN--CM, and HI--CM. This addresses the
observed trade-off: code-mixed adaptation improves EN--CM alignment
while degrading EN--HI alignment.

We introduce a trilingual post-training alignment stage that reduces the cosine distance between representations of parallel English, Hindi, and code-mixed sentences across all three language pairs.
Standard retrieval accuracy masks important asymmetries---for example, EN$\rightarrow$CM improving while  CM$\rightarrow$EN degrades. To measure balanced alignment, we introduce the Cross-Lingual Alignment Score (CLAS). CLAS accounts for bidirectional accuracy, directional bias, and variance across EN--CM, EN--HI, and HI--CM pairs.
Our main contributions are as follows:

\begin{itemize}
    \item We construct a trilingual corpus from existing resources comprising parallel EN, HI (Devanagari), and CM (Roman) sentences, enabling systematic analysis of cross-lingual alignment.

    \item Using this corpus, we analyze how multilingual encoders represent code-mixed text relative to their monolingual counterparts through CKA-based layer-wise similarity, token-level saliency, and entropy-based uncertainty analysis.

    \item Motivated by these insights, we introduce a trilingual post-training alignment stage that leverages parallel monolingual and code-mixed supervision to improve alignment across English, Hindi, and code-mixed representations. We further propose a metric, the Cross-Lingual Alignment Score (CLAS), to measure balanced cross-lingual alignment across language pairs.

    \item We evaluate the proposed alignment stage on downstream sentiment analysis and hate speech detection tasks for code-mixed text, demonstrating improved cross-lingual consistency performance.
\end{itemize}

\section{Related work}

\textbf{Multilingual Representation Learning.}
Multilingual language models, including mBERT~\cite{devlin-etal-2019-bert} and XLM-R~\cite{conneau-etal-2020-unsupervised}, learn shared representation spaces during multilingual pretraining, enabling effective cross-lingual transfer. Prior work shows that semantically similar sentences across languages tend to align in these spaces even without explicit supervision \cite{conneau-etal-2020-emerging, 10.1145/3764112}. This cross-lingual alignment ensures knowledge learned in one language benefits others in downstream tasks like NER and sentiment analysis, with methods like contrastive objectives further enhancing transfer \cite{kulshreshtha-etal-2020-cross}. While this property supports cross-lingual transfer, most studies \cite{hammerl-etal-2024-understanding} focus on monolingual text, leaving the behavior of these models under code-mixed inputs relatively underexplored.

\textbf{Code-mixed Setting.} Code-mixed language processing has received increasing attention in recent years \cite{sheth2025beyond}. This growing interest has led to the development of several datasets and benchmarks for Hindi–English and other code-mixed language pairs. Resources such as PHINC \cite{srivastava-singh-2020-phinc} and the LinCE benchmark \cite{aguilar-etal-2020-lince} provide parallel annotated corpora for studying code-mixing in social media text. Using these datasets, prior work has primarily focused on downstream tasks such as sentiment analysis, hate speech detection, and machine translation for code-mixed text \cite{niederreiter-gromann-2025-word, https://doi.org/10.1111/coin.70033, shanmugavadivel-etal-2024-code}. Recent studies also explore continued pretraining on large code-mixed corpora to improve task performance \cite{nayak-joshi-2022-l3cube, yoo-etal-2025-code-switching}. While these approaches improve task-level performance, they provide limited insight into how multilingual models internally represent code-mixed inputs.

\textbf{Constituent Language Influence.} Several prior works leverage monolingual language data to improve code-mixed sequence labeling and sequence classification tasks \cite{pant-dadu-2020-towards, jayanthi-etal-2021-codemixednlp, gautam-etal-2021-translate, kumar-etal-2022-utilizing, ogunremi-etal-2023-multilingual, 10.1145/3726866, mazumder-etal-2025-revealing}. These approaches consistently report performance gains on downstream tasks, demonstrating the practical utility of monolingual signals. Past studies also show that introducing code-mixing in in-context learning can improve performance, especially for low-resource languages \cite{shankar-etal-2024-context}, while decoder LLMs mix languages during their reasoning process \cite{wang-etal-2025-language-mixing}. This marks a broader synergy between code-mixed settings and their constituent languages.

\textbf{Interpretability.} Following the success of multilingual pretraining and fine-tuning approaches, subsequent works \cite{rogers-etal-2020-primer, resck-etal-2025-explainability} have investigated the internal mechanisms of multilingual language models using interpretability tools. In particular, representation similarity metrics such as CKA and SVCCA have been used to analyze how cross-lingual alignment emerges across transformer layers \cite{pmlr-v97-kornblith19a, wu-etal-2020-similarity}. Similar analyses have also examined how architectural factors and training dynamics influence representational alignment across independently trained networks \cite{10.5555/3327345.3327475, 10.5555/3295222.3295356}. Complementary studies \cite{banerjee2025attributional} explore attribution-based methods to understand token importance and language reliance within multilingual models.

Despite advances in multilingual representations, modeling and interpretability \cite{feng-etal-2022-language,10477442}, little work has examined how multilingual models internally represent code-mixed inputs \cite{santy-etal-2021-bertologicomix}. To our knowledge, existing studies do not systematically compare the representations of code-mixed inputs with those of their monolingual counterparts.
Our work addresses this gap by performing cross-lingual alignment evaluation driven by mechanistic interpretability analyses.

\section{Dataset}

\setcounter{footnote}{0}
We construct a unified trilingual corpus from existing resources by merging three publicly available datasets containing Hindi–English code-mixed text and their monolingual counterparts: CM-En Parallel \cite{dhar-etal-2018-enabling}, PHINC \cite{srivastava-singh-2020-phinc}, and the LinCE 2021 multiview Hinglish dataset\footnote{\url{https://github.com/devanshg27/cm_translation/tree/main}\label{fn:lince}} \cite{aguilar-etal-2020-lince, chen2022calcs}. These resources contain code-mixed sentences together with English (and/or Hindi) translations.

Since the first two datasets do not provide Hindi translations, we generate the missing Hindi sentences using the \texttt{Google Translate API}\footnote{\url{https://cloud.google.com/translate?hl=en}} and verify consistency of our results using the \texttt{IndicTrans2} model \cite{gala2023indictrans}, specialized English-to-Indic translation tool. After obtaining the translations, we unify all datasets into a trilingual corpus consisting of parallel English, Hindi (Devanagari), and Hindi–English code-mixed sentences.

Following prior work \cite{nayak-joshi-2022-l3cube}, we remove duplicate entries and filter out short sentences with fewer than five tokens to ensure sufficient code-mixing. The final dataset contains \textbf{21,139} aligned sentence triples. Each instance contains three fields: (i) \textbf{CM}, the Romanized Hindi–English code-mixed sentence, (ii) \textbf{English}, its monolingual English counterpart, and (iii) \textbf{Hindi}, the corresponding Hindi sentence written in Devanagari script. Additional dataset details are provided in Appendix~\ref{app:dataset_details}.

\begin{table}[t]
\centering
\small
\resizebox{\columnwidth}{!}{%
\begin{tabular}{|l|r|l|}
\hline
\textbf{Dataset} & \textbf{\# Samples} & \textbf{Languages} \\
\hline
CM-En parallel \cite{dhar-etal-2018-enabling} & 6,096 & CM, EN \\
PHINC \cite{srivastava-singh-2020-phinc} & 13,738 & CM, EN \\
LINCE 2021\footref{fn:lince}  & 8060 & CM, EN, HI \\ \hline
\end{tabular}}
\caption{Summary of the datasets used in this work, where EN = English, HI = Hindi (Devanagari script), and CM = Hindi (Roman-script) - English code-mixed text.}
\label{tab:dataset_statistics}
\end{table}

\begingroup
\renewcommand\thefootnote{}
\footnotetext{The terms \textit{alignment accuracy} and \textit{retrieval accuracy} are used interchangeably.}
\addtocounter{footnote}{-1}
\endgroup

\section{Experiments}
We design our experiments to systematically evaluate cross-lingual alignment in multilingual encoder language models using our trilingual corpus. Our observations are supported by mechanistic interpretability tools to connect observed behaviors to internal representation dynamics.

\paragraph{Models.}
We evaluate a set of encoder-based multilingual language models (MLMs), including general-purpose multilingual encoders trained on more than one hundred languages (mBERT and XLM-R) and their Hindi–English code-mixed adapted variants from the Hing family \cite{nayak-joshi-2022-l3cube} (see Table \ref{table:model_statistics}). This setup allows us to examine how different pretraining strategies influence cross-lingual alignment behavior.
We additionally evaluate IndicBERT-v2 and Qwen3-Embedding-0.6B to assess whether the observed alignment behavior extends beyond the mBERT and XLM-R families; the corresponding results are reported in Appendix~\ref{sec:additional_baselines}.

\begin{table}[h]
\centering
\setkeys{Gin}{keepaspectratio}
\resizebox{\columnwidth}{!}{%
\begin{tabular}{ll}
\hline
\textbf{Model} & \textbf{Version} \\
\hline
 mBERT                & \texttt{bert-base-multilingual-cased} \\
 Hing-mBERT           & \texttt{l3cube-pune/hing-bert} \\
 Hing-mBERT-Mixed   & \texttt{l3cube-pune/hing-bert-mixed} \\
 XLM-R            & \texttt{xlm-roberta-base} \\
 Hing-RoBERTa         & \texttt{l3cube-pune/hing-roberta} \\
 Hing-RoBERTa-Mixed & \texttt{l3cube-pune/hing-roberta-mixed} \\
 \hline
\end{tabular}}
\caption{Multilingual encoder models used in our experiments. Hing models denote mBERT and XLM-R variants further trained on the L3Cube-HingCorpus.}
\label{table:model_statistics}
\end{table}

\subsection{Cross-Lingual Alignment}
\label{sec:cross_lingual_alignment}
Here, we evaluate whether multilingual models learn structurally consistent representations between code-mixed text and its parallel monolingual counterparts. Structural consistency refers to the extent to which representations of parallel sentences occupy nearby regions in the latent space, independent of downstream task supervision. To assess this property, we perform representation similarity analysis across three language pairings: English--Hindi, English--code-mixed, and Hindi--code-mixed.

Our core evaluation method is a \textbf{dot product-based parallel sentence retrieval protocol}. For each sentence representation $\mathbf{h}_i^{\text{src}}$ in the source language, we compute dot products with a pool of candidate sentence representations in the target language consisting of one parallel (positive) sentence $\mathbf{h}_i^{\text{tgt}}$ and 10 sampled negative sentences $\{\mathbf{h}_j^{\text{tgt}}\}_{j \neq i}$ \cite{rahamim-belinkov-2024-contrasim}. Negative samples are drawn from a length-normalized candidate pool using percentile-based sentence length matching to avoid trivial mismatches. Formally,
\begin{equation}
    \text{score}(\mathbf{h}_i^{\text{src}}, \mathbf{h}_k^{\text{tgt}}) = \mathbf{h}_i^{\text{src}} \cdot \mathbf{h}_k^{\text{tgt}}
\end{equation}
We retrieve the sentence with the highest similarity score:
\begin{equation}
    \hat{k} = \argmax_{k \in \{i\} \cup \{j_1, \ldots, j_{10}\}} \text{score}(\mathbf{h}_i^{\text{src}}, \mathbf{h}_k^{\text{tgt}})
\end{equation}
Retrieval accuracy, defined as the proportion of queries for which $\hat{k}=i$, measures representational coherence across languages. To verify that results are not sensitive to the size of the negative pool, we additionally repeat the evaluation using 100 negatives. Details of the negative sample scaling, length-percentile sampling, and FAISS-based negative sampling procedure are provided in Appendix~\ref{app:negative_sampling}.

\begin{figure}[h]
    \centering
    \includegraphics[width=\columnwidth, keepaspectratio]{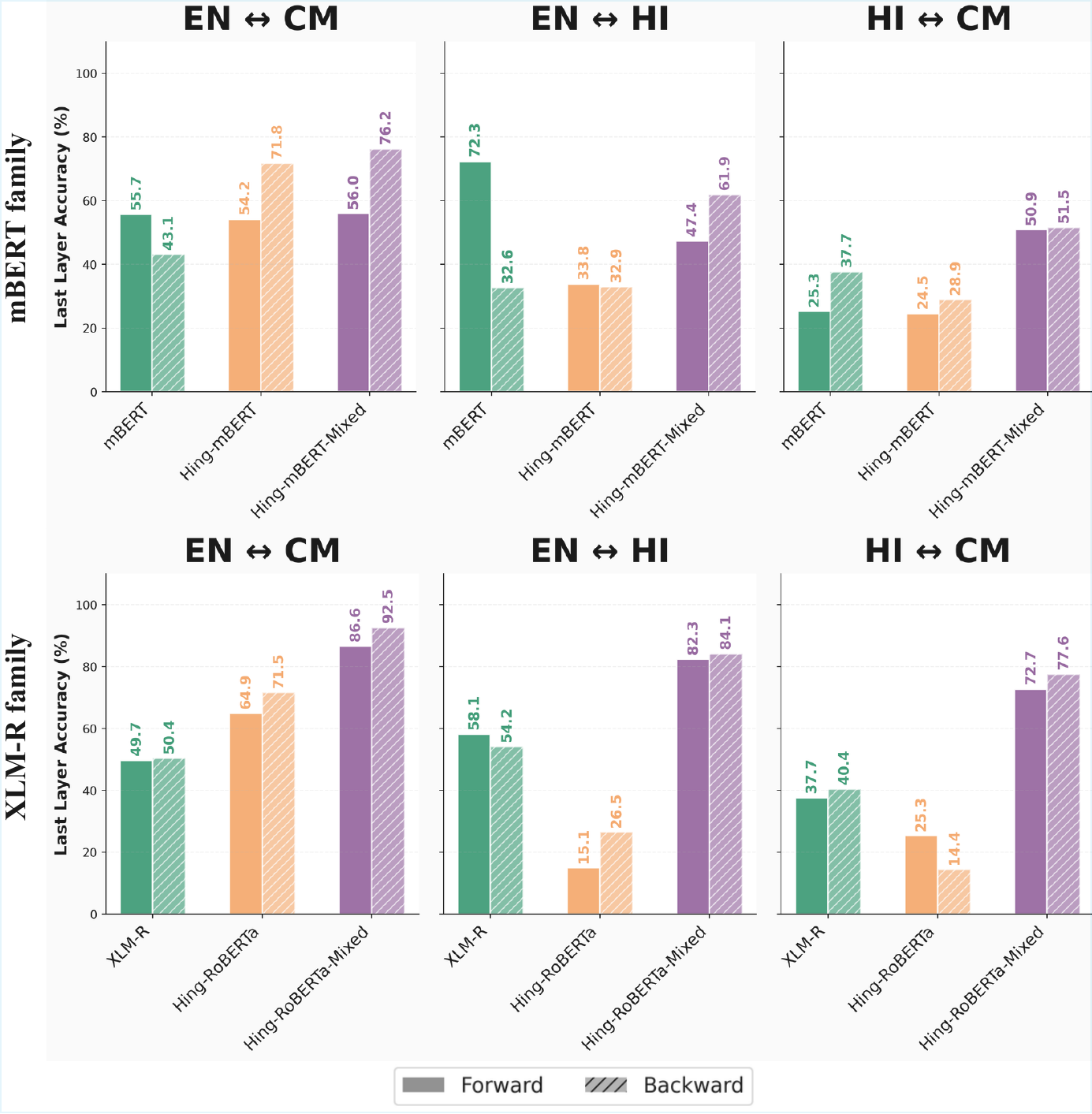}
    \caption{Directional cross-lingual alignment accuracy across model families. Solid bars show retrieval from language A$\rightarrow$B, hatched bars show B$\rightarrow$A.}
    \label{fig:crosslingual_ability_encoder_dot}
\end{figure}

\paragraph{Observations:} Here are our observations,
\begin{itemize}
    \item \textbf{Off-the-shelf model alignment.} Across base multilingual encoders, the EN–HI pair shows the strongest alignment (average of forward and backward) (see Figure~\ref{fig:crosslingual_ability_encoder_dot}). Code-mixed (CM) representations align weakly with both languages, yielding the ordering $\text{EN}\leftrightarrow\text{HI} > \text{EN}\leftrightarrow\text{CM} > \text{HI}\leftrightarrow\text{CM}$. This suggests that under standard multilingual pretraining, CM representations occupy a peripheral subspace that is weakly anchored to either monolingual language.

    \item \textbf{Trade-off between code-mixed and monolingual alignment.}
    Continued pretraining on Hinglish data produces a systematic trade-off: (i) EN$\leftrightarrow$CM alignment improves substantially (ii) while EN$\leftrightarrow$HI alignment degrades---a pair that the model had previously learned to align well through standard multilingual pretraining. After code-mixed adaptation, the alignment ordering inverts to: $\text{EN} \leftrightarrow \text{CM} > \text{EN} \leftrightarrow \text{HI} > \text{HI} \leftrightarrow \text{CM}$. This occurs because code-mixed data provides alignment signals primarily between English and code-mixed setting, causing English representations to be pulled toward the code-mixed manifold.
    These differences are primarily driven by pretraining data composition and script: code-mixed continued pretraining brings English and Romanized code-mixed representations closer through shared script and lexical overlap, while native-script Hindi remains comparatively distant.

    \item \textbf{Script-dependent modulation of Hindi--code-mixed alignment.} When Hindi is Romanized during pretraining, HI$\leftrightarrow$CM alignment degrades (Hing-mBERT: $\downarrow$23.33\%, Hing-RoBERTa: $\downarrow$64.3\% backward). In contrast, preserving Hindi in Devanagari improves alignment (Hing-mBERT-Mixed: $\uparrow$36.6\%, Hing-RoBERTa-Mixed: $\uparrow$92.0\%). These results suggest that script-level features influence representation geometry beyond lexical overlap, with Devanagari acting as a stronger cross-lingual anchor for Hindi--code-mixed alignment.

    \item \textbf{Reshaping anchor choice.} Beyond pretraining script choice, code-mixed adaptation also reshapes query anchor preference at evaluation time. In vanilla mBERT, the model favors \textit{Roman-script \& English-language} inputs as the dominant alignment source EN$\rightarrow$CM (55.7\%) and EN$\rightarrow$HI (72.3\%) substantially outperform their reverse directions. After Hinglish adaptation, this preference shifts toward \textit{Roman-script \& Hindi-language} anchors: the backward directions (CM$\rightarrow$EN: 76.2\%, HI$\rightarrow$CM: 51.5\%) now dominate, while forward English-as-source directions degrade.

    \item \textbf{Average layer-wise cross-lingual alignment.} When averaged across layers, base models (mBERT and XLM-R) show the highest alignment for EN$\leftrightarrow$HI, followed by EN$\leftrightarrow$CM and HI$\leftrightarrow$CM (Figure~\ref{fig:ret_acc_encoders_dot}; detailed scores in Appendix Table~\ref{tab:dot_percentile_layerwise}). After code-mixed adaptation, this ordering typically shifts to EN$\leftrightarrow$CM $>$ EN$\leftrightarrow$HI $>$ HI$\leftrightarrow$CM, reflecting a drop in EN$\leftrightarrow$HI alignment.

    The same pattern extends to modern multilingual baselines: IndicBERT-v2 and Qwen3-Embedding-0.6B obtain CLAS scores of 33.59 and 22.08, respectively, which increase to 51.47 and 79.59 after trilingual alignment.

\end{itemize}

\begin{findings}
    Code-mixed adaptation of multilingual encoders improve EN-CM alignment at the cost of EN-HI. It reorients the anchor choice for crosslingual alignment.
\end{findings}

\begin{figure}[h]
    \centering
    \includegraphics[width=\columnwidth, keepaspectratio]{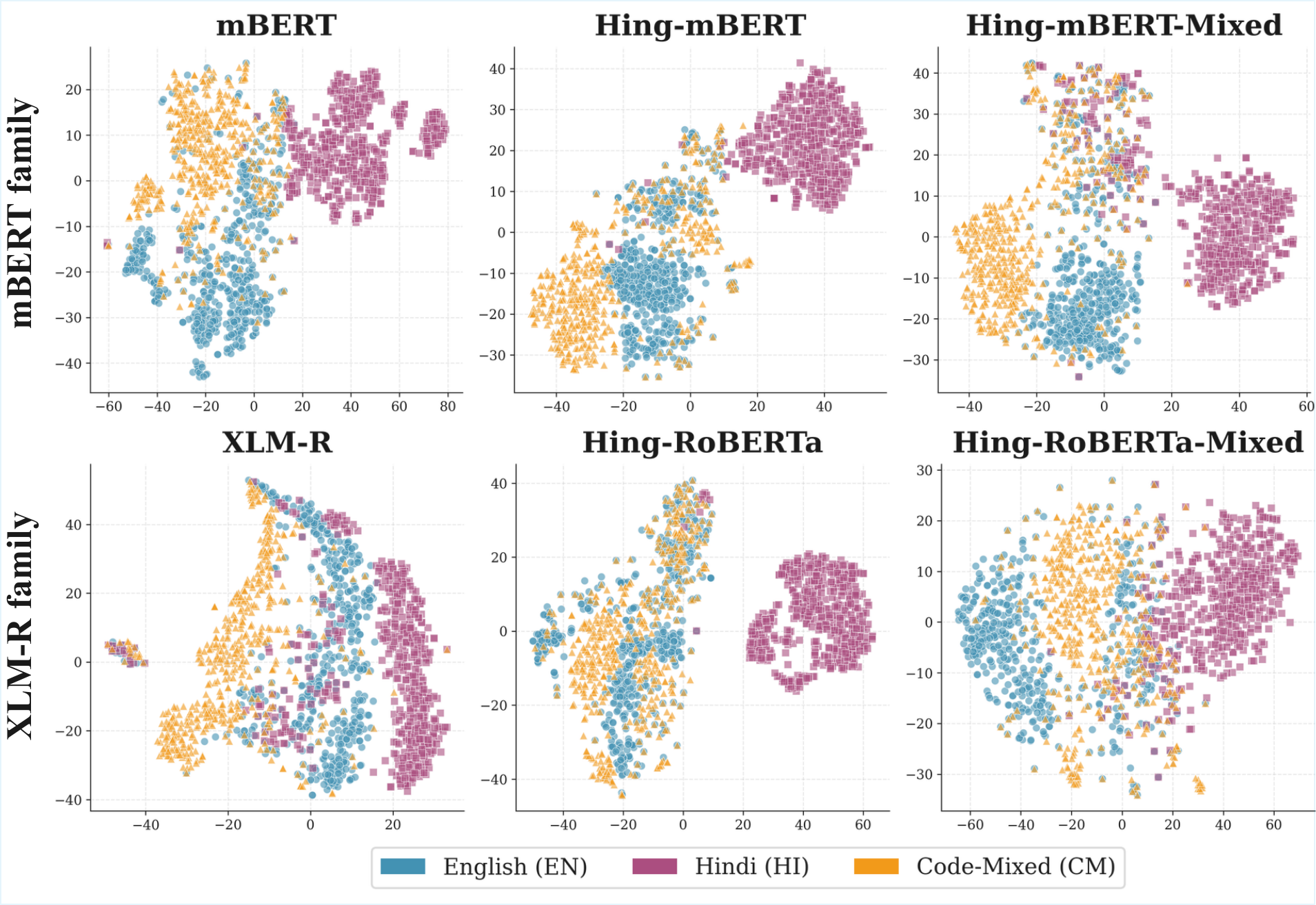}
    \caption{
    t-SNE visualization of sentence representations using mBERT and XLM-R-family models.
    }
    \label{fig:representation_Encoder_family_tsne}
\end{figure}

\begin{figure}[h]
    \centering
    \includegraphics[width=\columnwidth, keepaspectratio]{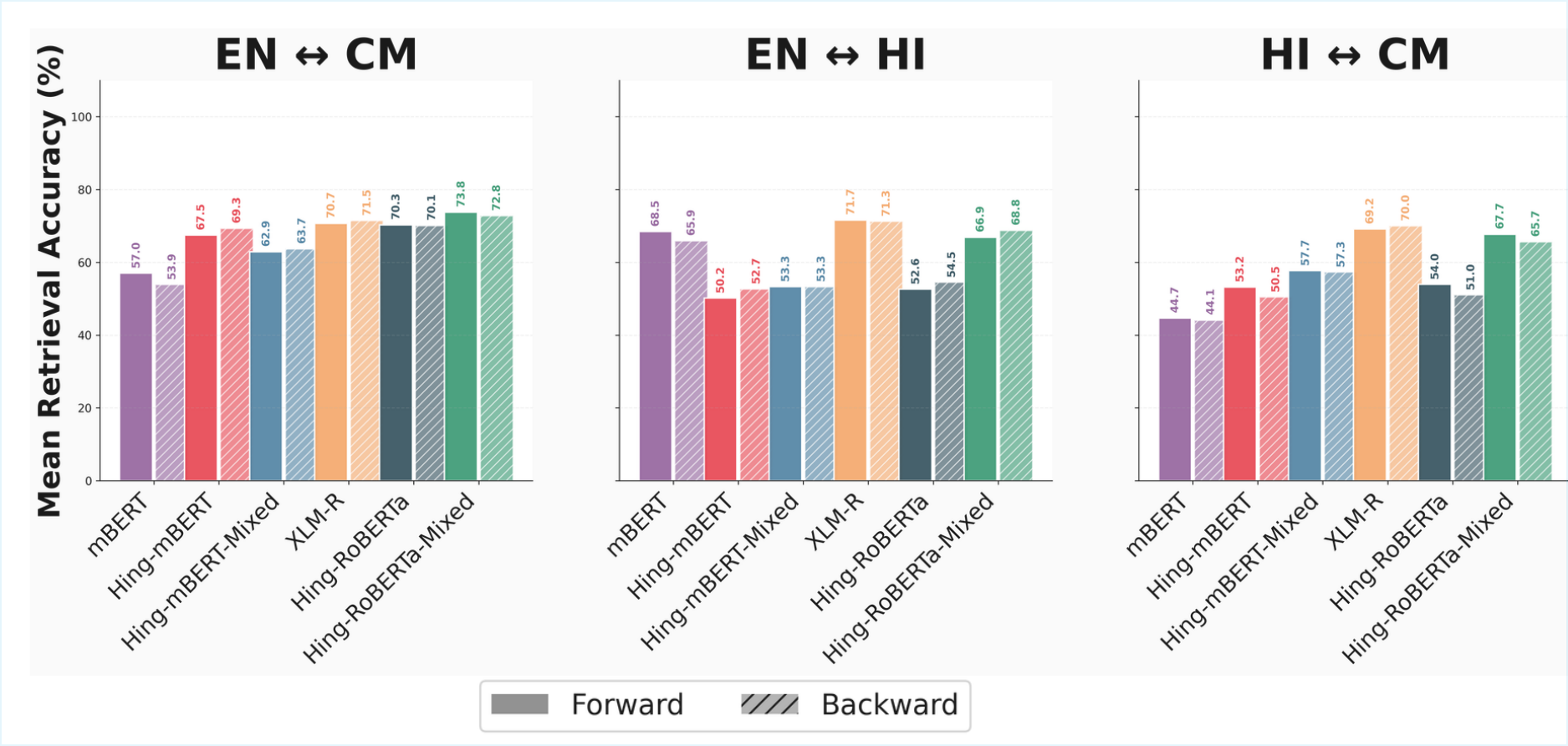}
    \caption{Directional dot-product retrieval accuracy across encoder models for EN–HI, EN–CM, and HI–CM, averaged across all layers. Solid bars denote A$\rightarrow$B and hatched bars denote B$\rightarrow$A.}

    \label{fig:ret_acc_encoders_dot}
\end{figure}

\begin{figure}[h]
    \centering
    \includegraphics[width=\columnwidth, keepaspectratio]{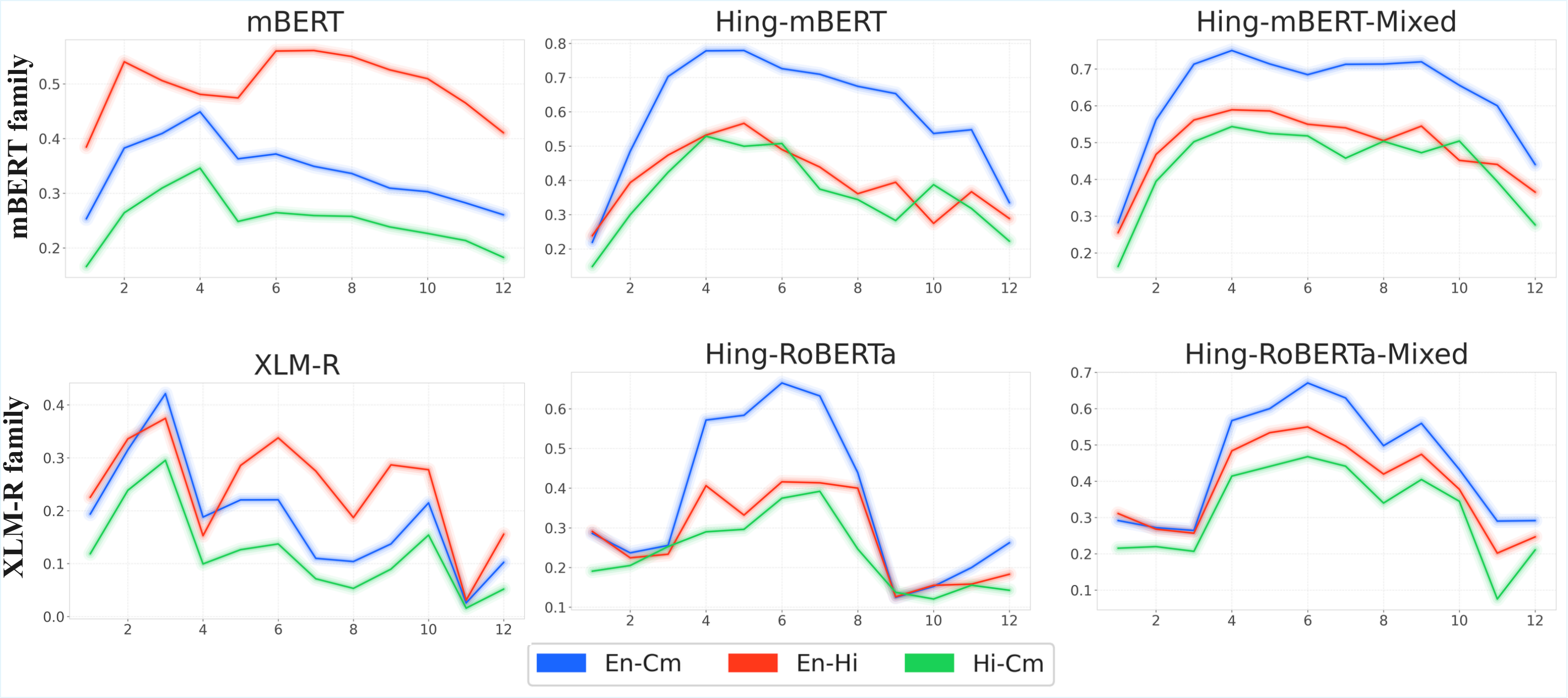}
    \caption{
        Layer-wise CKA alignment for mBERT and XLM-R family. Each subplot shows cross-lingual representation alignment for \textbf{EN-CM}, \textbf{EN-HI}, and \textbf{HI-CM} across layers.
    }
    \label{fig:cka_encoders_layerwise}
\end{figure}

\begin{figure}[t]
    \centering
    \includegraphics[width=\linewidth]{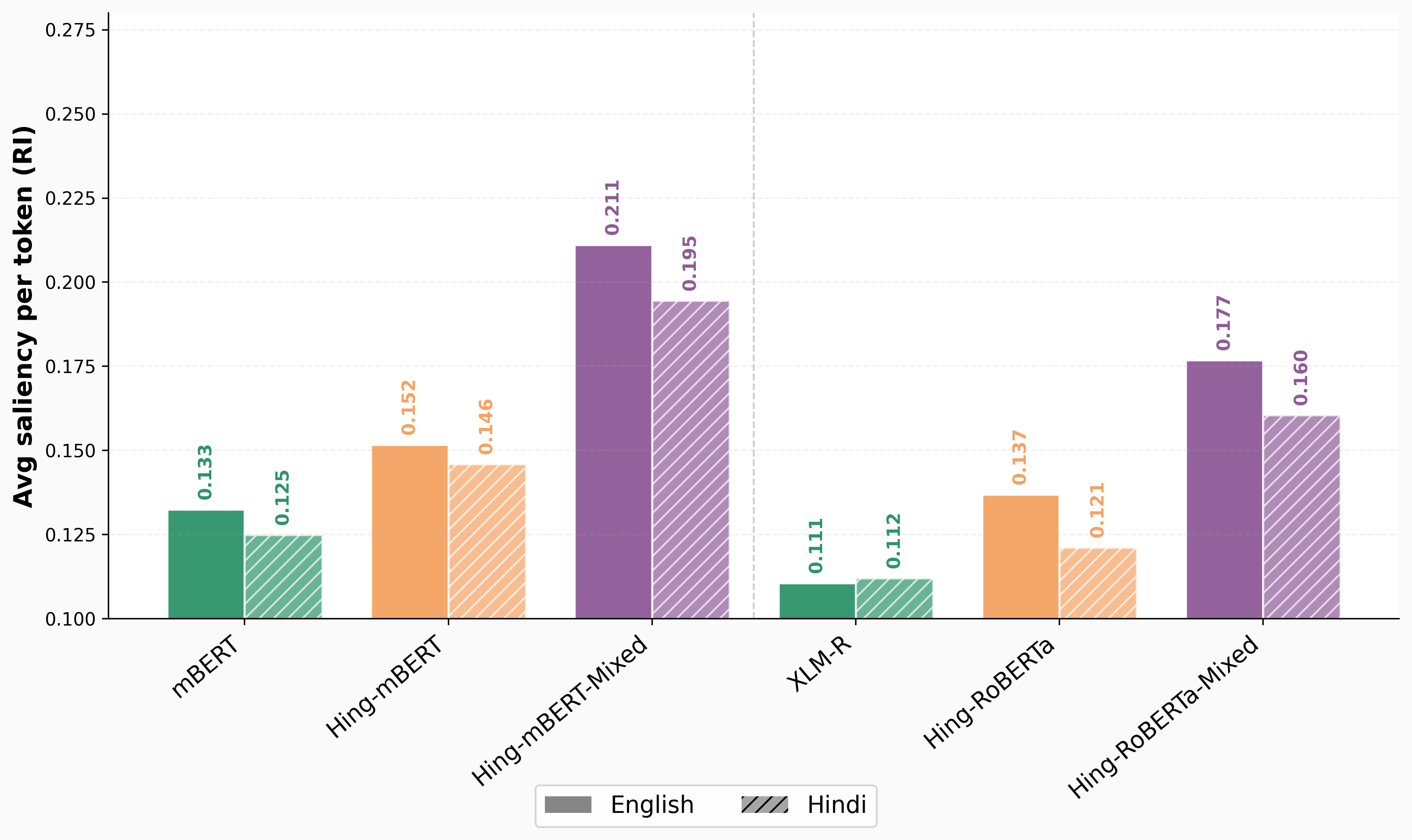}
    \caption{
    Language-wise per-token saliency for encoder-only models on Hinglish inputs.
    For each model, we report the average Relative Importance (RI) score assigned to English and Hindi tokens
    Special tokens (e.g., \texttt{<bos>}, \texttt{<eos>}) are excluded.    }
    \label{fig:encoderr-language-saliency}
\end{figure}

\subsection{Interpretability Analysis}
To better understand cross-lingual alignment among HI$\leftrightarrow$CM, EN$\leftrightarrow$CM, and EN$\leftrightarrow$HI, we analyze internal representations of MLMs across layers. Instead of relying only on final-layer embeddings, we examine how cross-lingual structure evolves across network layers. For this analysis, we use Centered Kernel Alignment (CKA), which measures similarity between representation spaces and correlates well with sentence-level alignment \cite{conneau-etal-2020-emerging}. This allows us to track how monolingual and code-mixed representations interact across layers.

\textbf{Layer-wise CKA evolution.}
Layer-wise CKA (Figure~\ref{fig:cka_encoders_layerwise}) reveals different alignment trajectories before and after code-mixed adaptation. In vanilla multilingual models (mBERT and XLM-R), CKA similarity typically peaks in the middle layers (around layers 3–6). In mBERT, EN–HI alignment dominates and peaks around layers 6–8 (CKA $\sim$0.56), while EN–CM and HI–CM similarities decline after the early layers (CKA $\sim$0.52 and $\sim$0.33). This partially contrasts with prior findings that cross-lingual similarity peaks in early layers \cite{conneau-etal-2020-emerging}, but aligns with \citet{liu-niehues-2025-middle}, who report middle-layer transfer behavior.

After continued pretraining on code-mixed data, the pattern changes. In Hing-mBERT, EN–CM similarity strengthens and peaks in the middle layers (CKA $\sim$0.78 around layers 4–5), while EN–HI similarity decreases. In the mixed-script variant (Hing-mBERT-Mixed), EN–CM similarity remains strong (CKA $\sim$0.75 at layer 4) and persists into deeper layers, while EN–HI pair remains comparatively weaker. Similar trends appear in the XLM-R family, where Hing-RoBERTa-Mixed maintains high similarity (CKA>$\sim$0.50) until approximately the 9\textsuperscript{th} layer. SVCCA analysis also shows similar trends (Figure~\ref{fig:svcca_encoders}). These results indicate that code-mixed pretraining shifts the representation space toward stronger EN–CM similarity.

\textbf{Language reliance in code-mixed understanding.}
We next analyze token-level language importance using the Relative Importance (RI) score \cite{banerjee2025attributional} (detailed in Appendix \ref{app:ri_analysis}). This analysis measures how much models rely on English versus Hindi tokens when processing code-mixed inputs. As shown in Figure~\ref{fig:encoderr-language-saliency}, all models assign substantially higher importance to English tokens. This suggests that code-mixed inputs are largely interpreted through an English-dominant semantic space, explaining the stronger EN–CM alignment compared to HI–CM.

\textbf{Geometric organization of multilingual representations.}
We further examine representation geometry using t-SNE projections (Figure~\ref{fig:representation_Encoder_family_tsne}). In vanilla mBERT, EN, HI, and CM form clearly separated clusters. After code-mixed adaptation, EN and CM representations collapse into a shared region, while Hindi remains relatively distant. Similar patterns appear in XLM-R and Hing-RoBERTa. However, the strongest alignment scores (e.g., Hing-RoBERTa-Mixed with 86.6\% EN$\leftrightarrow$CM and 72.7\% HI$\leftrightarrow$CM retrieval) correspond to a different geometry where CM representations lie between EN and HI with partial overlap. This suggests that effective alignment emerges when CM acts as a bridge between EN and HI.

\textbf{Entropy-based uncertainty reduction.}
Finally, we analyze how monolingual signals explain code-mixed representations using entropy-based uncertainty reduction. For each layer, we compute the entropy of CM representations and the reduction in entropy when conditioned on English, Hindi, or both languages (details in Appendix~\ref{app:entropy_analysis}). Larger reductions indicate stronger explanatory power. Across models, conditioning on English consistently yields greater uncertainty reduction than conditioning on Hindi, indicating that CM representations are more strongly explained by English. However, conditioning jointly on both languages produces the largest reduction. This suggests that Hindi still provides complementary information for understanding CM inputs.

These findings highlight the importance of native-script information. Hindi written in Devanagari preserves orthographic and lexical cues that are often lost in Romanized CM text, helping anchor code-mixed representations in the multilingual space.

\begin{figure}[htbp]
\centering
\includegraphics[width=\columnwidth]{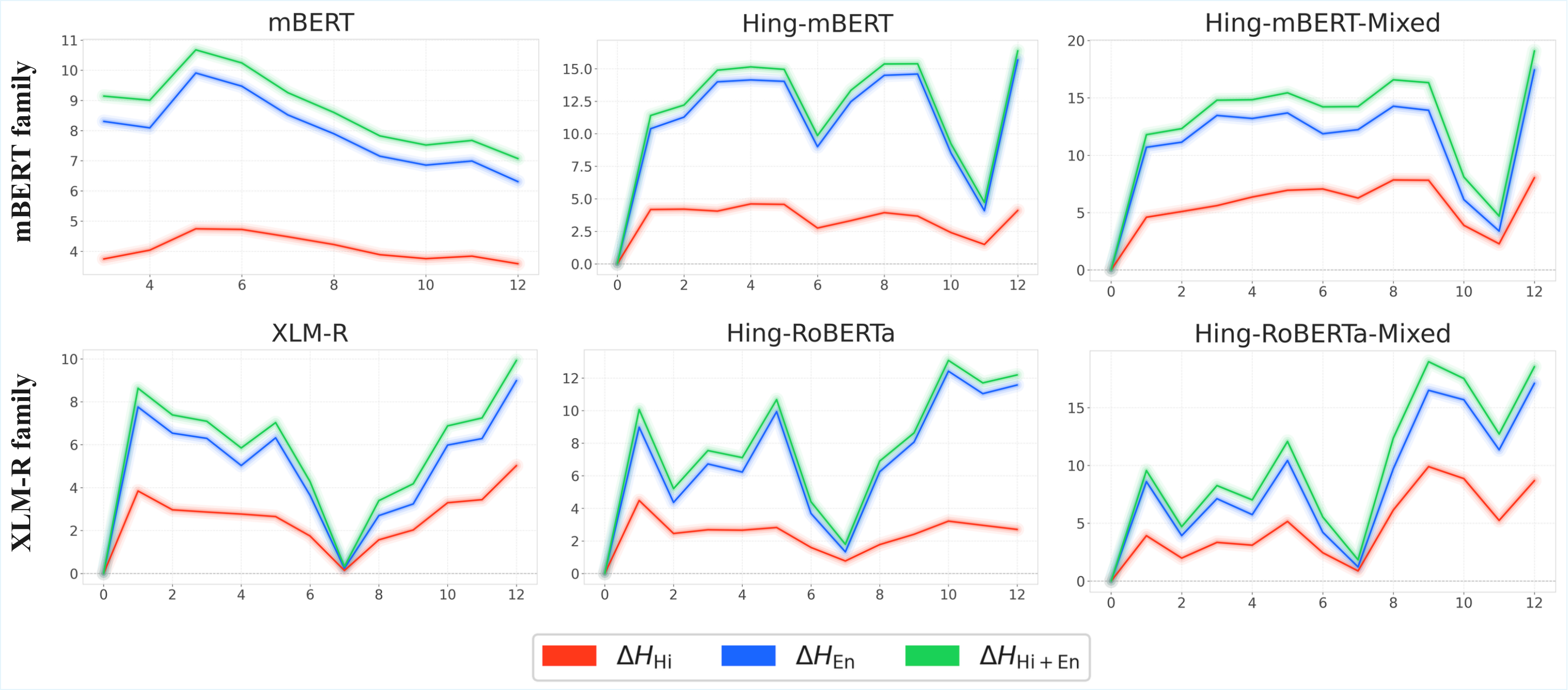}
\caption{Uncertainty reduction in code-mixed (CM) representations across encoder layers in MLMs.
}
\label{fig:entropy_delta_aligned_models}
\end{figure}

\begin{findings}
    Code-mixed representations are largely aligned with English in multilingual models, while native-script Hindi provides complementary signals that help reduce uncertainty.
\end{findings}

\subsection{Our proposed: Trilingual Post-Training Alignment Stage}

Let $\mathcal{D} = \{(x^{en}_i, x^{hi}_i, x^{cm}_i)\}_{i=1}^{N}$ be a trilingual corpus of aligned sentence triples in English, Hindi, and romanized code-mixed. The corpus is split into 80\% training and 20\% test data for post-training and alignment evaluation. We introduce a trilingual post-training alignment stage that adapts pretrained multilingual encoders using the objective:
\begin{equation}
    \mathcal{L} = \mathcal{L}_{\text{base}} + \lambda \cdot \mathcal{L}_{\text{align}}
\end{equation}
where $\lambda = 0.05$ and $\mathcal{L}_{\text{base}}$ denotes the original pretraining objective of the underlying encoder. For mBERT, this includes the Masked Language Modeling (MLM) and Next Sentence Prediction (NSP) losses \cite{devlin-etal-2019-bert}, while for XLM-R it corresponds to the MLM objective alone \cite{conneau-etal-2020-unsupervised}.

For architectures with a sentence-level objective (e.g., NSP in mBERT), we apply it to cross-lingual sentence pairs from $(x^{en}, x^{hi})$, $(x^{en}, x^{cm})$, and $(x^{hi}, x^{cm})$, treating translation pairs as positives and randomly sampled sentences as negatives. Models without such objectives (e.g., XLM-R) are trained using MLM together with the proposed alignment loss.

\paragraph{Cross-Lingual alignment loss.}
The key component of our method is $\mathcal{L}_{\text{align}}$,which reduces the cosine distance between representations of semantically equivalent sentences across EN, HI, and CM. For each example $i$ in a batch of size $B$, we obtain $\ell_2$-normalized \texttt{[CLS]} embeddings:
\begin{equation}
\begin{aligned}
    \mathbf{e}_i &= \frac{f_\theta(x^{en}_i)}{\|f_\theta(x^{en}_i)\|}, \quad
    \mathbf{h}_i = \frac{f_\theta(x^{hi}_i)}{\|f_\theta(x^{hi}_i)\|}, \\
    \mathbf{c}_i &= \frac{f_\theta(x^{cm}_i)}{\|f_\theta(x^{cm}_i)\|}
\end{aligned}
\end{equation}
where $f_\theta(\cdot)$ is the encoder. The alignment loss is the mean
cosine distance across all three language pairs:
\begin{equation}
\begin{aligned}
\mathcal{L}_{\text{align}} &=
\frac{1}{3} \sum_{(u,v)}
\frac{1}{B}\sum_{i=1}^{B}
\left(1 - \mathbf{u}_i^{\top}\mathbf{v}_i\right), \\
& where \quad (u,v) \in \{(e,h),(e,c),(h,c)\}.
\end{aligned}
\end{equation}

Note that the alignment objective does not explicitly optimize directional retrieval symmetry or variance across language pairs. Instead, by jointly aligning EN, HI, and CM sentence representations, the loss encourages a shared embedding geometry that implicitly promotes more balanced cross-lingual alignment across language pairs. Rest of the training details are provided in Appendix~\ref{app:training_details}, and the pseudocode for mBERT training is shown in Algorithm~\ref{alg:pretraining}.

\paragraph{Cross-Lingual alignment evaluation.}
Beyond forward and backward accuracies, we evaluate whether the model learns a \emph{balanced} cross-lingual representation space across the three setups: EN$\leftrightarrow$CM, EN$\leftrightarrow$HI, and HI$\leftrightarrow$CM. Let $A_{xy}^{\rightarrow}$ and $A_{xy}^{\leftarrow}$ denote forward and backward alignment accuracies between languages $x$ and $y$.
We define the \textsc{Cross-Lingual Alignment Score (CLAS)} as
\[
\text{CLAS} = \text{MeanAcc} - \text{DirBias} - \text{SetupStd}.
\]

Here, $\text{MeanAcc} = \frac{1}{6}\sum_{(x,y)}(A_{xy}^{\rightarrow}+A_{xy}^{\leftarrow})$ measures the average bidirectional retrieval accuracy across language pairs,
$\text{DirBias} = \frac{1}{3}\sum_{(x,y)}|A_{xy}^{\rightarrow}-A_{xy}^{\leftarrow}|$ captures directional asymmetry between languages, and
$\text{SetupStd} = \mathrm{Std}\!\left(\frac{A_{xy}^{\rightarrow}+A_{xy}^{\leftarrow}}{2}\right)$ measures performance variance across the three cross-lingual setups.
Since each accuracy term lies in $[0,100]$, CLAS is bounded in the range $[-150,100]$. Higher values indicate strong bidirectional alignment with low directional bias and balanced performance across language pairs.

\paragraph{Observations}

\begin{table}[htbp]
\centering
\resizebox{\columnwidth}{!}{%
\begin{tabular}{@{}lccccccc@{}}
\toprule
& \multicolumn{2}{c}{\textbf{EN$\leftrightarrow$CM}} & \multicolumn{2}{c}{\textbf{EN$\leftrightarrow$HI}} & \multicolumn{2}{c}{\textbf{HI$\leftrightarrow$CM}} & \textbf{CLAS} \\
\cmidrule(lr){2-3} \cmidrule(lr){4-5} \cmidrule(lr){6-7}
\textbf{Model} & $\rightarrow$ & $\leftarrow$ & $\rightarrow$ & $\leftarrow$ & $\rightarrow$ & $\leftarrow$ \\
\midrule
\multicolumn{8}{l}{\textit{Last Layer Retrieval}} \\
\midrule
\multicolumn{8}{l}{\textbf{mBERT Family}} \\
\quad mBERT & 54.75 & 42.90 & 74.87 & 33.80 & 26.30 & 39.05 & 14.20  \\
\quad mBERT Trilingual & 69.43 & 63.49 & 71.52 & 73.23 & 54.49 & 49.69 & \textbf{50.97} \\
\quad Hing-mBERT & 52.78 & 71.86 & 32.30 & 33.43 & 24.70 & 28.99 & 17.01 \\
\quad Hing-mBERT Trilingual  & 57.73 & 62.96 & 78.05 & 76.40 & 53.00 & 59.70 & \textbf{51.07} \\
\quad Hing-mBERT-Mixed & 54.26 & 74.63 & 41.81 & 60.73 & 49.51 & 48.94 & 34.95 \\
\quad Hing-mBERT-Mixed Trilingual & 54.64 & 62.42 & 71.52 & 70.06 & 41.96 & 49.15 & \textbf{42.51} \\
\midrule
\multicolumn{8}{l}{\textbf{XLM-R Family}} \\
\quad XLM-R & 50.56 & 50.67 & 58.81 & 54.33 & 38.37 & 40.46 & 39.53 \\
\quad XLM-R Trilingual & 56.06 & 54.18 & 53.43 & 54.50 & 48.37 & 46.93 & \textbf{47.50} \\
\quad Hing-RoBERTa & 64.94 & 72.14 & 15.02 & 27.54 & 26.19 & 14.48 & 3.74  \\
\quad Hing-RoBERTa Trilingual & 81.43 & 89.85 & 69.11 & 76.70 & 63.84 & 62.58 & \textbf{58.98}\\
\quad Hing-RoBERTa-Mixed & 86.95 & 92.37 & 83.20 & 84.98 & 72.88 & 77.78 & 73.09  \\
\quad Hing-RoBERTa-Mixed Trilingual & 86.40 & 93.31 & 84.65 & 84.08 & 73.27 & 80.96 & \textbf{73.50}  \\
\midrule
\midrule
\multicolumn{8}{l}{\textit{Mean Retrieval}} \\
\midrule
\multicolumn{8}{l}{\textbf{mBERT Family}} \\
\quad mBERT & 58.77 & 57.51 & 59.35 & 60.15 & 45.10 & 42.73 & 45.35 \\
\quad mBERT Trilingual & 65.15 & 63.47 & 71.05 & 70.55 & 48.26 & 49.20 & \textbf{50.98}\\
\quad Hing-mBERT & 65.76 & 68.57 & 44.43 & 49.58 & 50.64 & 46.15 & 40.84 \\
\quad Hing-mBERT Trilingual  & 64.87 & 65.90 & 58.17 & 60.18 & 60.86 & 59.98 & \textbf{57.67} \\
\quad Hing-mBERT-Mixed & 60.57 & 62.31 & 48.89 & 50.63 & 56.25 & 55.72 & 49.62 \\
\quad Hing-mBERT-Mixed Trilingual  & 66.78 & 67.82 & 56.11 & 60.56 & 58.01 & 55.44 & \textbf{53.45} \\
\midrule
\multicolumn{8}{l}{\textbf{XLM-R Family}} \\
\quad XLM-R & 67.77 & 68.34 & 68.61 & 68.24 & 66.24 & 67.04 & 66.36\\
\quad XLM-R Trilingual & 72.26 & 72.16 & 71.96 & 72.11 & 71.07 & 70.84 & \textbf{71.02}\\
\quad Hing-RoBERTa & 69.40 & 68.93 & 54.33 & 56.81 & 55.75 & 52.45 & 50.75\\
\quad Hing-RoBERTa Trilingual & 71.62 & 73.85 & 67.55 & 69.61 & 64.36 & 64.21 & \textbf{63.60}\\
\quad Hing-RoBERTa-Mixed  & 72.05 & 70.72 & 64.73 & 66.92 & 65.41 & 63.02 & 62.10 \\
\quad Hing-RoBERTa-Mixed Trilingual & 70.41 & 68.01 & 65.57 & 65.87 & 65.68 & 62.90 & \textbf{62.51}  \\
\bottomrule
\end{tabular}%
}
\caption{Bidirectional cross-lingual alignment accuracy (\%) using dot-product similarity with 10 negative samples evaluated on the test split. All scores are average of three random seeds.}
\label{tab:retrieval_bidirectional_dotproduct_10_negatives}
\end{table}

The crosslingual alignment performances are presented in Table \ref{tab:retrieval_bidirectional_dotproduct_10_negatives}. Here, we present our observations:

\begin{itemize}
    \item Our proposed method yields large improvements in CLAS, particularly for mBERT ($14.20 \rightarrow 50.97$) and XLM-R ($39.53 \rightarrow 47.50$), indicating reduced directional bias and more balanced alignment across the three language pairs. The alignment objective also mitigates the degradation of EN$\leftrightarrow$HI alignment observed after code-mixed adaptation.

    \item After code-mixed adaptation, preserving cross-lingual capability becomes crucial; our method improves the Hing models by restoring balanced alignment across languages.

    \item Similar trends hold when the negative sample pool is increased to 100 (Table~\ref{tab:retrieval_bidirectional_dotproduct_100_neg}). A variant trained only with parallel supervision (without the alignment loss) also performs competitively (Table~\ref{tab:retrieval_dotproduct_ablation}).
\end{itemize}

From an interpretability perspective, Figure \ref{fig:cka_trilingual_alignment_layerwise} shows layer-wise CKA alignment where the language pairs exhibit closer and more consistently high alignment across layers.
We additionally compare the proposed alignment objective with InfoNCE and AnInfoNCE on the same trilingual sentence pairs. Both alternatives improve cross-lingual alignment, while the proposed objective achieves comparable mean CLAS and downstream consistency; detailed results are reported in Appendix~\ref{sec:alternative_objectives}.

\paragraph{Downstream task validation.}
To examine whether improved alignment translates to practical benefits, we evaluate the proposed post-training alignment stage on two code-mixed classification tasks: sentiment analysis and hate speech detection. Detailed experimental settings and results are provided in Appendix~\ref{app:downstream_tasks}. Overall, the aligned models, in general show improved cross-lingual consistency scores after trilingual alignment, as summarized by the mean consistency scores in Table~\ref{tab:consistency}. This trend holds across both tasks and multiple model families, suggesting that trilingual alignment encourages stable multilingual representations and improves cross-lingual robustness.

\begin{table}[t]
\centering
\setlength{\tabcolsep}{3pt}
\begin{tabular}{lcc}
\toprule
\textbf{Model} & \textbf{Sentiment} & \textbf{Hate} \\
\midrule
mBERT     & 0.4012 & 0.5220 \\
mBERT-Tri & \textbf{0.4898} & \textbf{0.5953} \\
XLM-R     & 0.5720 & 0.6307 \\
XLM-R-Tri & \textbf{0.5986} & \textbf{0.6701} \\
\bottomrule
\end{tabular}
\caption{Mean \textit{consistency} scores across CM, EN, and HI training settings for sentiment analysis and hate speech detection. The definition of \textit{consistency} metric and additional scores are presented in Appendix~\ref{app:downstream_tasks}.}

\label{tab:consistency}
\end{table}

\begin{findings}
    Trilingual supervision improves cross-lingual alignment across EN–HI–CM and leads to better cross-lingual performance on downstream tasks.
\end{findings}

\section{Conclusion}
In this work, we studied how multilingual encoders represent Hindi–English code-mixed text and how these representations relate to their constituent languages. Our analysis shows that in multilingual language models (MLMs), code-mixed representations often lie on the periphery of the representation space and are weakly anchored to both languages. Continued adaptation on code-mixed data improves alignment between English and code-mixed inputs but can degrade alignment between the original monolingual languages. Our interpretability analyses further reveal that multilingual models tend to interpret code-mixed text through an English-dominant semantic space, while native-script Hindi provides complementary signals that help reduce representational uncertainty. Motivated by these observations, we introduce a simple yet effective trilingual post-training alignment stage that encourages consistent representations across English, Hindi, and code-mixed inputs. Experiments demonstrate improved cross-lingual alignment while reducing degradation in EN–HI alignment. As a result, downstream sentiment and hate speech tasks also show improved cross-lingual consistency.

\section*{Limitations}

This study focuses on Hindi--English code-mixing due to our linguistic expertise. While this pair provides a useful case study, the extent to which the observed mechanisms generalize to other code-mixed settings remains an open question. Settings involving multiple scripts, intra-word mixing, or more than two languages may exhibit different representational dynamics. Future work can evaluate these mechanisms across structurally diverse code-mixed language pairs and settings.

Our post-training alignment stage uses a relatively small trilingual corpus (21K sentence triples). Although the results show improved alignment, larger and more diverse code-mixed corpora may lead to stronger and more stable gains. Future work could scale the alignment stage with larger datasets or explore data augmentation strategies for generating additional code-mixed samples.

Our study focuses on encoder-based multilingual models (e.g., mBERT, XLM-R). While these models remain widely used for multilingual representation learning, extending the proposed alignment approach to other encoders and larger decoder-based multilingual LLMs remains an open direction for future work.

Another limitation arises from the use of automatic translations when constructing the trilingual corpus and preparing certain evaluation data. Although machine translation systems provide reasonable approximations, they may not fully preserve the linguistic nuances, stylistic variations, or pragmatic cues present in the original text. This issue is particularly relevant for code-mixed language, where subtle interactions between languages may be altered during translation. Future work could address this limitation by incorporating human-verified translations or collecting larger trilingual corpora directly from multilingual speakers.

\section*{Ethical considerations}
This work studies social media text and may contain potentially offensive or harmful language present in the original datasets. These instances are included solely for research purposes in order to study linguistic phenomena in code-mixed settings, and the authors do not endorse or promote any harmful or abusive content. AI-based writing assistants were used for grammar checking and language polishing during manuscript preparation.

\bibliography{custom}

\appendix
\appendix

\section*{Appendix}

\section{Additional Dataset Details}
\label{app:dataset_details}

\subsection{CM-En parallel \cite{dhar-etal-2018-enabling}}
We used the English–Hindi code-mixed parallel corpus introduced by \citet{dhar-etal-2018-enabling}. The dataset contained 6,096 English–Hindi code-mixed sentences paired with gold-standard English translations, which are produced by fluent bilingual annotators and have achieved high inter-annotator agreement (Fleiss’ $\kappa \approx 0.88$). The sentences are written in Roman script and reflect informal social communication. The corpus showed a Code-Mixing Index (CMI) of 30.5, indicating a substantial degree of code-mixing.

\subsection{PHINC: Parallel Hinglish Social Media Corpus \cite{srivastava-singh-2020-phinc}}

The second dataset we integrated was the PHINC corpus \cite{srivastava-singh-2020-phinc}. The dataset contained 13,738 Hindi–English code-mixed sentences, each paired with a parallel human-written English translation. The sentences are collected from social media platforms such as Twitter (now X) and Facebook. Thus similar to the previous one, the dataset reflected an informal style of writing. The corpus is curated through extensive filtering and manually annotated by 54 expert-level annotators, with each annotator labeling 400 randomly selected samples. The code-mixed variant here exhibited a relatively higher degree of code-mixing with CMI being 75.76.

\subsection{LinCE Multiview Hinglish Dataset \cite{aguilar-etal-2020-lince}}

We also used the Hindi–English code-mixed dataset from the LinCE benchmark\footref{fn:lince} introduced by \citet{aguilar-etal-2020-lince, chen2022calcs}.
It contained 8060 samples crawled from wikipedia article about a particular movie. This subset is compiled by consolidating and re-curating existing Hindi–English code-mixed corpora collected from social media platforms such as Twitter (now X) and Facebook. Each instance included (i) an English sentence, (ii) its Hindi translation in Devanagari script, (iii) a Hindi (Roman-script) - English code-mixed form, and (iv) a Hindi (Devanagari-script) - English code-mixed variant along with token-level language tags. Here, the code-mixed corpus showed a CMI of 22.68.

\subsection{Translation Quality and Translator Validation}
\label{app:translation_validation}

To assess the quality of the Hindi translations used to construct the trilingual corpus, we evaluate the Google Translate outputs through English backtranslation. Table~\ref{tab:backtranslation_quality} reports these translation quality scores and they are acceptable as per prior art \cite{bhattacharjee-etal-2024-domain}.

\begin{table}[h]
\centering
\small
\begin{tabular}{lc}
\hline
Metric & Score \\
\hline
Average BLEU & 0.2527 \\
Average SacreBLEU & 38.73 \\
Average COMET & 0.8347 \\
\hline
\end{tabular}
\caption{Backtranslation-based evaluation of the Hindi translations used in the trilingual corpus.}
\label{tab:backtranslation_quality}
\end{table}

We further assess whether the choice of translation system affects the observed alignment patterns by reconstructing the Hindi translations using IndicTrans2 \cite{gala2023indictrans}. Table~\ref{tab:translator_validation} compares the resulting alignment scores with those obtained using Google Translate. The scores remain comparable across the two systems, indicating that the observed alignment patterns are not sensitive to the choice of translator.

\begin{table*}[t]
\centering
\small
\begin{tabular}{lccccccc}
\hline
 & EN-CM & CM-EN & EN-HI & HI-EN & HI-CM & CM-HI & CLAS \\
\hline
\multicolumn{8}{l}{\textbf{Last layer}}\\
mBERT (Google Translate) & 54.75 & 42.90 & 74.87 & 33.80 & 26.30 & 39.05 & 14.20 \\
XLM-R (Google Translate) & 50.56 & 50.67 & 58.81 & 54.33 & 38.37 & 40.46 & \textbf{39.53} \\
mBERT (IndicTrans2) & 55.37 & 42.98 & 72.76 & 32.46 & 25.35 & 37.79 & 13.53 \\
XLM-R (IndicTrans2) & 50.15 & 50.55 & 58.34 & 54.01 & 38.17 & 40.44 & \textbf{39.28} \\
\hline
\multicolumn{8}{l}{\textbf{Mean layer}}\\
mBERT (Google Translate) & 58.77 & 57.51 & 59.35 & 60.15 & 45.10 & 42.73 & 45.34 \\
XLM-R (Google Translate) & 67.77 & 68.34 & 68.61 & 68.24 & 66.24 & 67.04 & \textbf{66.36} \\
mBERT (IndicTrans2) & 53.32 & 49.98 & 66.00 & 63.12 & 39.93 & 39.42 & 39.56 \\
XLM-R (IndicTrans2) & 68.36 & 69.07 & 69.27 & 68.84 & 66.67 & 67.63 & \textbf{66.78} \\
\hline
\end{tabular}
\caption{Comparison of cross-lingual alignment scores obtained using Hindi translations generated by Google Translate and IndicTrans2.}
\label{tab:translator_validation}
\end{table*}

\section{Trilingual Training Details}
\label{app:training_details}

We post-train the base multilingual encoders (mBERT and XLM-R) using the proposed trilingual alignment objective on the curated English--Hindi--code-mixed corpus. The dataset is split into 80\% training and 20\% test data.
Models are trained for up to 10 epochs using the AdamW optimizer with a batch size of 16. We explore learning rates from the set $\{5e\!-\!6,\,6e\!-\!6,\,6.5e\!-\!6,\,1e\!-\!5\}$ and apply a linear warmup schedule over the first 10\% of the total training steps. During training, we follow the standard BERT masked language modeling (MLM) protocol with 15\% token masking, where 80\% of the selected tokens are replaced with \texttt{[MASK]}, 10\% with random tokens, and 10\% are left unchanged.
The alignment loss weight $\lambda$ is tuned over $\{0.05,\,0.10,\,0.15\}$. Each configuration is trained with three random seeds, and we report the average performance across runs. Although training loss continues to decrease across epochs, improvements in cross-lingual alignment become marginal in later stages. Based on alignment performance on the validation split, we select the 6\textsuperscript{th} epoch for XLM-R and the 8\textsuperscript{th} epoch for mBERT for final evaluation.

\section{Additional Results}
\label{app:additional_results}

\begin{figure*}[h]
    \centering
    \includegraphics[width=\textwidth, keepaspectratio]{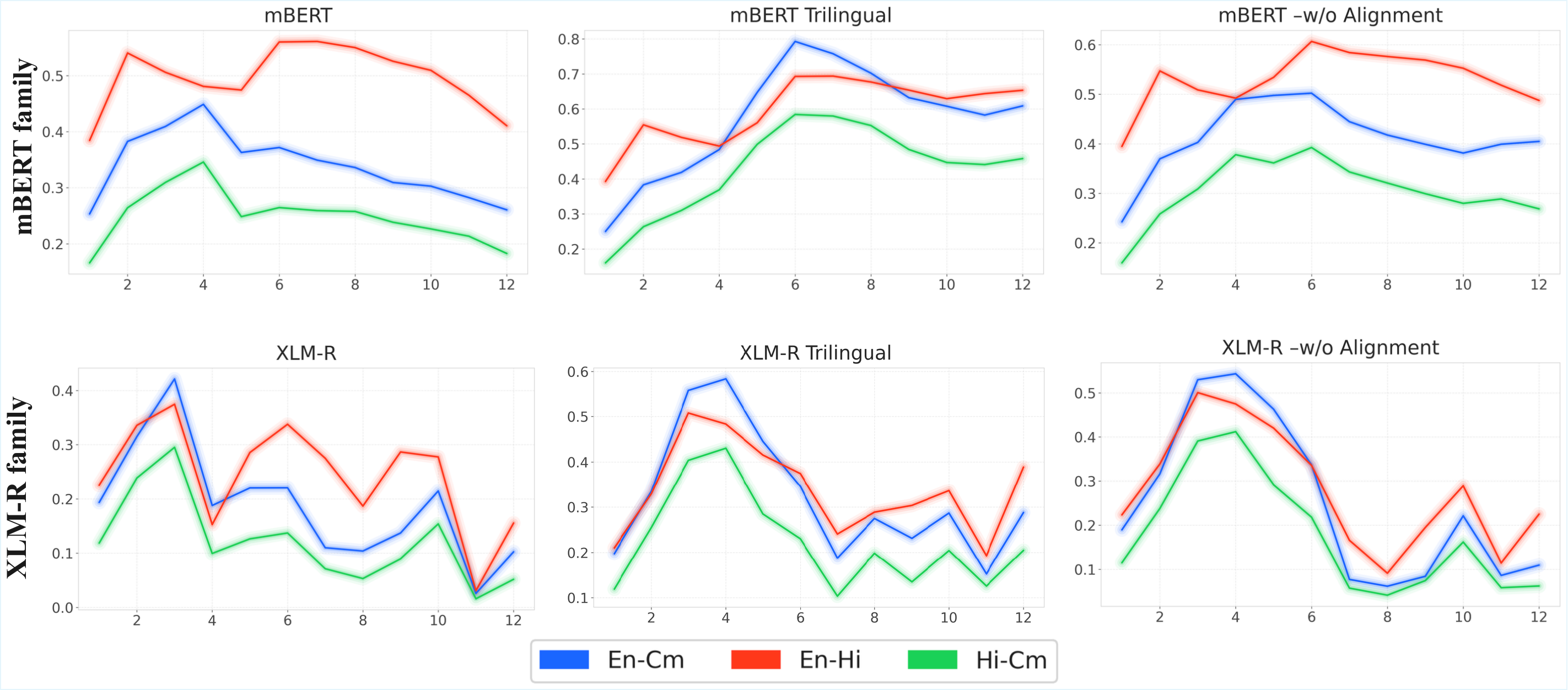}
    \caption{
        Layer-wise CKA alignment for encoders and their trilingual versions. Each subplot shows cross-lingual representation alignment for \textbf{EN-CM}, \textbf{EN-HI}, and \textbf{HI-CM} across layers.
    }
    \label{fig:cka_trilingual_alignment_layerwise}
\end{figure*}

This section presents additional results supporting our analysis. We include an ablation study in which the alignment loss is removed from the training objective to examine its impact on cross-lingual alignment. Table~\ref{tab:retrieval_dotproduct_ablation} reports the corresponding CLAS values. Even without the alignment loss, training with parallel trilingual supervision alone yields slightly better cross-lingual alignment compared to the base models, in most of the cases. We also report perplexity scores on the code-mixed test set in Table~\ref{tab:ppl_cm}. The perplexity of the trilingual models remains lower than, or comparable to, that of the base models, indicating that the alignment objective does not adversely affect language modeling performance on code-mixed inputs. Finally, Figure~\ref{fig:cka_trilingual_alignment_layerwise} shows the layer-wise CKA representations of the model trained without the alignment loss.

\subsection{Additional Modern Baselines}
\label{sec:additional_baselines}

We additionally evaluate IndicBERT-v2 and Qwen3-Embedding-0.6B to assess whether the observed alignment behavior extends beyond the mBERT and XLM-R families (see Table \ref{tab:additional_baselines} for exact versions). As shown in Tables~\ref{tab:alignment} and \ref{tab:indicbert_consistency} , both models exhibit weak CM alignment in their pretrained form, while trilingual alignment substantially improves CLAS and consistently improves cross-lingual consistency on the downstream tasks. These results further support the model-agnostic nature of the observed alignment behavior.

\begin{table}[t]
\centering
\resizebox{\columnwidth}{!}{%
\begin{tabular}{ll}
\hline
\textbf{Model} & \textbf{Version} \\
\hline
IndicBERT-v2 & \texttt{ai4bharat/IndicBERTv2-MLM-only} \\
Qwen3-Embedding & \texttt{Qwen/Qwen3-Embedding-0.6B} \\
\hline
\end{tabular}%
}
\caption{Additional multilingual encoder models used in our experiments.}
\label{tab:additional_baselines}
\end{table}

\begin{table}[t]
\centering
\small
\begin{tabular}{lcc}
\toprule
\textbf{Model} & \textbf{Last Layer} & \textbf{Mean} \\
\midrule
IndicBERTv2        & 33.59 & 34.14 \\
IndicBERTv2 Tri    & \textbf{51.47} & \textbf{37.27} \\
Qwen3-Embedding    & 22.08 & 50.25 \\
Qwen3-Embedding Tri & \textbf{79.59} & \textbf{67.70} \\
\bottomrule
\end{tabular}
\caption{Dot-product alignment scores for IndicBERT-v2, its trilingual variant, Qwen3-Embedding-0.6B, and its trilingual variant. Reported values correspond to CLAS.}
\label{tab:alignment}
\end{table}

\begin{table}[t]
\centering
\resizebox{\columnwidth}{!}{%
\begin{tabular}{@{}llccc@{}}
\toprule
\textbf{Task} & \textbf{Model} & \textbf{Train CM} & \textbf{Train EN} & \textbf{Train HI} \\
\midrule
\multicolumn{5}{l}{\textbf{Hate Speech Detection}} \\
\midrule
\quad & IndicBERTv2 & 0.6340 & 0.6097 & 0.6142 \\
\quad & IndicBERTv2 Tri & \textbf{0.6576} & \textbf{0.6438} & \textbf{0.6496} \\
\midrule
\multicolumn{5}{l}{\textbf{Sentiment Analysis}} \\
\midrule
\quad & IndicBERTv2 & 0.6346 & 0.4932 & 0.6063 \\
\quad & IndicBERTv2 Tri & \textbf{0.6393} & \textbf{0.5014} & \textbf{0.6154} \\
\bottomrule
\end{tabular}%
}
\caption{Cross-lingual consistency scores for IndicBERT-v2 on hate speech detection and sentiment analysis.}
\label{tab:indicbert_consistency}
\end{table}

\subsection{Comparison with Alternative Alignment Objectives}
\label{sec:alternative_objectives}

To further assess the effectiveness of the proposed trilingual alignment objective, we compare it with two standard contrastive alignment objectives, InfoNCE~\cite{oord2019representationlearningcontrastivepredictive} and AnInfoNCE~\cite{rusak2025infonceidentifyinggaptheory}, using the same parallel trilingual data. Table~\ref{tab:infonce_alignment} reports the resulting cross-lingual alignment scores, while Tables~\ref{tab:infonce_hate} and~\ref{tab:infonce_sentiment} compare their downstream cross-lingual consistency on hate speech detection and sentiment analysis, respectively.
Across both mBERT and XLM-R, all evaluated alignment objectives improve cross-lingual alignment over the pretrained baselines. InfoNCE and AnInfoNCE often achieve higher CLAS than the proposed objective, particularly at the last layer, while mean CLAS scores are broadly comparable across objectives. However, higher alignment scores do not consistently translate into stronger downstream consistency, indicating that downstream gains are not determined solely by the choice of alignment loss or the magnitude of CLAS.

\begin{table}[htbp]
\centering
\resizebox{\columnwidth}{!}{%
\begin{tabular}{@{}llcc@{}}
\toprule
\textbf{Model Family} & \textbf{Training Objective} & \textbf{Last Layer} & \textbf{Mean} \\
\midrule
\multicolumn{4}{l}{\textit{CLAS }} \\
\midrule
\multicolumn{4}{l}{\textbf{mBERT Family}} \\
\quad mBERT & Standard & 14.20 & 66.36 \\
\quad mBERT & Tri & 50.97 & 71.02 \\
\quad mBERT & InfoNCE & \textbf{86.78} & 70.07 \\
\quad mBERT & AnInfoNCE & 83.79 & \textbf{73.20} \\
\midrule
\multicolumn{4}{l}{\textbf{XLM-R Family}} \\
\quad XLM-R & Standard & 39.53 & 66.19 \\
\quad XLM-R & Tri & 47.50 & 72.26 \\
\quad XLM-R & InfoNCE & \textbf{58.14} & 71.15 \\
\quad XLM-R & AnInfoNCE & 53.97 & \textbf{73.21} \\
\bottomrule
\end{tabular}%
}
\caption{Comparison of alignment objectives using dot-product retrieval. Reported values correspond to CLAS.}
\label{tab:infonce_alignment}
\end{table}

\begin{table}[t]
\centering
\small
\begin{tabular}{lccc}
\toprule
Model & Train CM & Train EN & Train HI \\
\midrule
mBERT & 0.5542 & 0.4805 & 0.5314 \\
mBERT Tri & 0.6194 & \textbf{0.6119} & 0.5545 \\
mBERT InfoNCE & \textbf{0.6207} & 0.5627 & 0.5903 \\
mBERT AnInfoNCE & 0.5971 & 0.5700 & \textbf{0.6100} \\
\midrule
XLM-R & 0.6217 & 0.6326 & 0.6379 \\
XLM-R-Tri & \textbf{0.6794} & \textbf{0.6732} & 0.6577 \\
XLM-R InfoNCE & 0.6348 & 0.6593 & 0.6495 \\
XLM-R AnInfoNCE & 0.6493 & 0.6504 & \textbf{0.6648} \\
\bottomrule
\end{tabular}
\caption{Cross-lingual consistency scores for hate speech detection under alternative alignment objectives.}
\label{tab:infonce_hate}
\end{table}

\begin{table}[t]
\centering
\small
\begin{tabular}{lccc}
\toprule
Model & Train CM & Train EN & Train HI \\
\midrule
mBERT & 0.3283 & 0.3674 & 0.5079 \\
mBERT Tri & 0.4464 & \textbf{0.4710} & 0.5519 \\
mBERT InfoNCE & 0.5433 & 0.4657 & \textbf{0.5760} \\
mBERT AnInfoNCE & \textbf{0.5553} & 0.4562 & 0.5550 \\
\midrule
XLM-R & 0.6304 & 0.4686 & 0.6170 \\
XLM-R-Tri & \textbf{0.6618} & 0.5022 & 0.6318 \\
XLM-R InfoNCE & 0.6616 & \textbf{0.5333} & \textbf{0.6464} \\
XLM-R AnInfoNCE & 0.6407 & 0.5162 & 0.6306 \\
\bottomrule
\end{tabular}
\caption{Cross-lingual consistency scores for sentiment analysis under alternative alignment objectives.}
\label{tab:infonce_sentiment}
\end{table}

\begin{table}[htbp]
\centering
\resizebox{\columnwidth}{!}{%
\begin{tabular}{@{}lccccccc@{}}
\toprule
& \multicolumn{2}{c}{\textbf{EN$\leftrightarrow$CM}} & \multicolumn{2}{c}{\textbf{EN$\leftrightarrow$HI}} & \multicolumn{2}{c}{\textbf{HI$\leftrightarrow$CM}} & \textbf{CLAS} \\
\cmidrule(lr){2-3} \cmidrule(lr){4-5} \cmidrule(lr){6-7}
\textbf{Model} & $\rightarrow$ & $\leftarrow$ & $\rightarrow$ & $\leftarrow$ & $\rightarrow$ & $\leftarrow$ \\
\midrule
\multicolumn{7}{l}{\textit{Last Layer Retrieval}} \\
\midrule
\multicolumn{7}{l}{\textbf{mBERT Family}} \\
\quad mBERT & 54.75 & 42.90 & 74.87 & 33.80 & 26.30 & 39.05 & 14.20 \\
\quad mBERT Trilingual & \textbf{69.43} & \textbf{63.49} & 71.52 & \textbf{73.23} & \textbf{54.49} & 49.69 & \textbf{50.97} \\
\quad \quad --w/o Alignment & 52.90 & 47.49 & 66.83 & 30.40 & 22.91 & 36.55 & 15.06 \\
\midrule
\multicolumn{7}{l}{\textbf{XLM-R Family}} \\
\quad XLM-R & 50.56 & 50.67 & 58.81 & 54.33 & 38.37 & 40.46 & 39.53 \\
\quad XLM-R Trilingual & \textbf{56.06} & \textbf{54.18} & 53.43 & 54.50 & \textbf{48.37} & \textbf{46.93} & \textbf{47.50}  \\
\quad \quad --w/o Alignment & 55.92 & 54.23 & 63.94 & 60.96 & 42.92 & 43.14 & 43.88 \\
\midrule
\midrule
\multicolumn{7}{l}{\textit{Mean Retrieval }} \\
\midrule
\multicolumn{7}{l}{\textbf{mBERT Family}} \\
\quad mBERT & 58.60 & 57.42 & 59.37 & 60.10 & 44.99 & 42.65 & 45.35 \\
\quad mBERT Trilingual & \textbf{65.15} & \textbf{63.47} & \textbf{71.05} & \textbf{70.55} & 48.26 & 49.20 & \textbf{50.98} \\
\quad \quad --w/o Alignment& 55.94 & 53.24 & 68.33 & 63.32 & 42.74 & 43.10 & 42.40 \\
\midrule
\multicolumn{7}{l}{\textbf{XLM-R Family}} \\
\quad XLM-R & 67.77 & 68.34 & 68.61 & 68.24 & 66.24 & 67.04 & 66.36 \\
\quad XLM-R Trilingual & \textbf{72.26} & \textbf{72.16} & \textbf{71.96} & \textbf{72.11} & \textbf{71.07} & \textbf{70.84} & \textbf{71.02} \\
\quad \quad --w/o Alignment & 72.62 & 71.50 & 73.06 & 72.30 & 71.09 & 70.62 & 70.32 \\
\bottomrule
\end{tabular}%
}
\caption{Bidirectional alignment scores using dot product similarity. Last layer accuracy (\%), mean accuracy (\%) and CLAS (\%) are reported for model families evaluated on the test split. All scores are average of three random seeds.}
\label{tab:retrieval_dotproduct_ablation}
\end{table}

\begin{table}[htbp]
\centering
\resizebox{\columnwidth}{!}{%
\begin{tabular}{@{}lccccccc@{}}
\toprule
& \multicolumn{2}{c}{\textbf{EN$\leftrightarrow$CM}} & \multicolumn{2}{c}{\textbf{EN$\leftrightarrow$HI}} & \multicolumn{2}{c}{\textbf{HI$\leftrightarrow$CM}} & \textbf{CLAS} \\
\cmidrule(lr){2-3} \cmidrule(lr){4-5} \cmidrule(lr){6-7}
\textbf{Model} & $\rightarrow$ & $\leftarrow$ & $\rightarrow$ & $\leftarrow$ & $\rightarrow$ & $\leftarrow$ \\
\midrule
\multicolumn{8}{l}{\textit{Last Layer Retrieval}} \\
\midrule
\multicolumn{8}{l}{\textbf{mBERT Family}} \\
\quad mBERT & 36.46 & 28.11 & 52.07 & 16.41 & 11.66 & 19.06 & 1.68 \\
\quad mBERT Trilingual & 51.73 & 46.61 & 57.66 & 60.80 & 36.02 & 31.57 & \textbf{32.70} \\
\quad Hing-mBERT & 35.11 & 46.36 & 26.85 & 14.70 & 12.03 & 10.72 & 3.82 \\
\quad Hing-mBERT Trilingual & 27.70 & 42.55 & 53.19 & 51.80 & 16.84 & 24.92 & \textbf{15.10} \\
\quad Hing-mBERT-Mixed &38.86 & 57.58 & 23.51 & 43.53 & 32.36 & 29.23 & 15.91 \\
\quad Hing-mBERT-Mixed Trilingual & 31.91 & 42.67 & 62.98 & 62.13 & 30.13 & 39.83 &\textbf{ 25.35} \\
\midrule
\multicolumn{8}{l}{\textbf{XLM-R Family}} \\
\quad XLM-R & 30.62 & 30.27 & 36.73 & 34.23 & 19.76 & 20.65 & 21.11 \\
\quad XLM-R Trilingual & 47.72 & 45.17 & 46.19 & 47.07 & 40.28 & 38.60 & \textbf{39.12} \\
\quad Hing-RoBERTa & 50.43 & 57.43 & 3.64 & 10.10 & 9.09 & 3.35 & -6.39 \\
\quad Hing-RoBERTa Trilingual & 67.55 & 78.90 & 50.54 & 61.75 & 44.68 & 44.25 & \textbf{38.47} \\
\quad Hing-RoBERTa-Mixed & 72.58 & 81.60 & 71.84 & 74.72 & 58.15 & 63.56 & 57.70 \\
\quad Hing-RoBERTa-Mixed Trilingual & 85.43 & 93.35 & 86.07 & 83.99 & 71.33 & 82.71 & \textbf{71.56} \\
\midrule
\midrule
\multicolumn{8}{l}{\textit{Mean Retrieval}} \\
\midrule
\multicolumn{8}{l}{\textbf{mBERT Family}} \\
\quad mBERT & 34.70 & 32.39 & 44.49 & 44.24 & 22.74 & 20.85 & 22.54 \\
\quad mBERT Trilingual & 49.43 & 48.10 & 55.46 & 55.24 & 38.55 & 38.10 & \textbf{39.81} \\
\quad Hing-mBERT & 50.28 & 52.38 & 29.70 & 34.26 & 35.12 & 31.00 & 26.32 \\
\quad Hing-mBERT Trilingual & 51.18 & 53.00 & 40.22 & 43.82 & 41.10 & 38.96 & \textbf{36.91} \\
\quad Hing-mBERT-Mixed & 42.61 & 45.22 & 30.04 & 33.87 & 38.80 & 36.85 & 30.22 \\
\quad Hing-mBERT-Mixed Trilingual & 47.02 & 48.99 & 40.45 & 42.75 & 43.21 & 42.33 & \textbf{39.62} \\
\midrule
\multicolumn{8}{l}{\textbf{XLM-R Family}} \\
\quad XLM-R & 56.97 & 58.12 & 57.11 & 57.05 & 55.53 & 56.49 & 55.51\\
\quad XLM-R Trilingual & 63.37 & 63.41 & 63.06 & 63.50 & 62.57 & 62.24 & \textbf{62.32}\\
\quad Hing-RoBERTa & 53.69 & 56.13 & 36.62 & 38.38 & 35.28 & 35.65 & 32.28 \\
\quad Hing-RoBERTa Trilingual & 60.57 & 62.91 & 57.25 & 58.17 & 51.59 & 53.74 & \textbf{51.86} \\
\quad Hing-RoBERTa-Mixed & 59.95 & 59.37 & 52.45 & 54.55 & 53.61 & 51.50 & 50.49 \\
\quad Hing-RoBERTa-Mixed Trilingual & 70.26 & 68.66 & 64.61 & 66.77 & 66.28 & 62.84 & \textbf{62.08} \\
\bottomrule
\end{tabular}%
}
\caption{Bidirectional cross-lingual alignment accuracy (\%) using dot-product similarity with 100 negative samples evaluated on the test split. All scores are average of three random seeds.}
\label{tab:retrieval_bidirectional_dotproduct_100_neg}
\end{table}

\begin{table}[h]
\centering
\resizebox{\columnwidth}{!}{%
\begin{tabular}{llc}
\toprule
\textbf{Model Family} & \textbf{Model} & \textbf{PPL (CM)} \\
\midrule
\multirow{6}{*}{mBERT}
    & mBERT                           & 462.41 \\
    & mBERT Trilingual                  & 119.28 \\
    & Hing-mBERT                           & 10.98 \\
    & Hing-mBERT Trilingual              & 8.08 \\
    & Hing-mBERT-Mixed                     & 10.40 \\
    & Hing-mBERT-Mixed Trilingual        & 8.37 \\
\midrule
\multirow{6}{*}{XLM-R}
    & XLM-R                           & 52.54 \\
    & XLM-R Trilingual                    & 24.85 \\
    & Hing-RoBERTa                         & 16.19 \\
    & Hing-RoBERTa Trilingual             &  9.19 \\
    & Hing-RoBERTa-Mixed                   & 16.41 \\
    & Hing-RoBERTa-Mixed Trilingual      & 13.97\\
\bottomrule
\end{tabular}%
}
\caption{Perplexity scores computed on the code-mixed test set.}
\label{tab:ppl_cm}
\end{table}

\section{Negative Sample Scaling and Sampling Strategies}
\label{app:negative_sampling}

This section provides additional implementation details for the cross-lingual retrieval evaluation described in Section~\ref{sec:cross_lingual_alignment}. In particular, we describe how candidate pools are constructed for the retrieval task and the negative sampling strategies used to generate challenging distractors. For each query sentence representation in the source language, the candidate set consists of the correct parallel sentence (positive sample) together with a set of negative sentences drawn from the target language corpus. Unless otherwise stated, the main experiments use ten negative samples per query. We further analyze the robustness of the retrieval protocol by scaling the number of negatives to 100 and by evaluating different sampling strategies for constructing the negative pool.

\subsection{Scaling negative samples}
\label{sec:scaling_neg_samples}
To examine whether our findings are sensitive to the size of the negative candidate pool in the retrieval protocol, we repeat the cross-lingual alignment evaluation using 100 randomly sampled negative sentences for each query instead of 10. This increases the difficulty of the retrieval task and provides a stricter test of representational alignment. The resulting alignment accuracies are presented in Table~\ref{tab:retrieval_bidirectional_dotproduct_100_neg}. Overall, the trends remain consistent with our earlier results, indicating that our observations are robust to larger negative pools.

\subsection{Length-Aware Percentile Sampling}

The first strategy constructs candidate pools using a length-aware sampling procedure. For each sentence in the dataset, we compute the token length and determine its percentile rank within the corpus. Given a query sentence with percentile rank $p$, candidate negatives are restricted to sentences whose percentile ranks fall within a window of $\pm5$ percentiles around $p$. This constraint ensures that negative examples have comparable sentence lengths, reducing trivial cues that could make retrieval artificially easy.

From this candidate pool, ten negative sentences are sampled uniformly at random while excluding the true parallel sentence. This approach creates retrieval settings where negative examples are length-matched but otherwise randomly selected.

Layerwise retrieval accuracies obtained using this sampling strategy are reported in Table~\ref{tab:dot_percentile_layerwise}.

\subsection{FAISS-Based Hard Negative Sampling}

To construct more challenging retrieval scenarios, we additionally employ a FAISS\cite{8733051} based negative sampling strategy. In this setting, candidate pools are first constructed using the same length-aware percentile filtering described above. However, instead of sampling negatives uniformly, we use the final-layer sentence representations to guide the sampling process.

Specifically, cosine similarities between the query representation and all candidate sentences are computed using a FAISS inner-product index. Negative examples are then sampled probabilistically based on these similarity scores, with lower-similarity sentences receiving higher sampling probability. This procedure increases the likelihood of selecting semantically similar distractors while still avoiding trivial matches.

Layerwise retrieval results obtained using FAISS-based sampling are reported in Table~\ref{tab:dot_faiss_layerwise}. These results provide a more stringent evaluation of cross-lingual alignment by introducing harder negative examples.

\begin{algorithm}[h]
\caption{Trilingual Post-training Alignment of mBERT}
\label{alg:pretraining}
\begin{algorithmic}[1]

\State \textbf{Input:} Trilingual dataset $\mathcal{D}=\{(x^{en}_i,x^{hi}_i,x^{cm}_i)\}_{i=1}^{N}$, pretrained encoder $f_\theta$, hyperparameters $\lambda,T$
\State \textbf{Output:} Trilingual-aligned encoder $f_\theta$

\For{epoch $=1$ to $T$}
    \For{mini-batch $\mathcal{B}\subset\mathcal{D}$}

        \Statex \textbf{MLM \& NSP:}
        \State Sample sentence pair from $\{(en,hi),(en,cm),(hi,cm)\}$
        \State Construct positive/negative NSP pair and apply token masking
        \State Compute $\mathcal{L}_{MLM}$ and $\mathcal{L}_{NSP}$ using $f_\theta$

        \Statex \textbf{Cross-lingual alignment:}
        \State Encode each language independently:
        \State $\mathbf{e}_i \leftarrow \ell_2\text{-normalize}(f_\theta(x^{en}_i)[\texttt{CLS}])$
        \State $\mathbf{h}_i \leftarrow \ell_2\text{-normalize}(f_\theta(x^{hi}_i)[\texttt{CLS}])$
        \State $\mathbf{c}_i \leftarrow \ell_2\text{-normalize}(f_\theta(x^{cm}_i)[\texttt{CLS}])$

        \State $\displaystyle
        \mathcal{L}_{align} =
        \frac{1}{3B}\sum_{i=1}^{B} \Big[
        (1-\mathbf{e}_i^\top\mathbf{h}_i)
        + (1-\mathbf{e}_i^\top\mathbf{c}_i)
        + (1-\mathbf{h}_i^\top\mathbf{c}_i)
        \Big]$

        \State $\mathcal{L} \leftarrow \mathcal{L}_{MLM} + \mathcal{L}_{NSP} + \lambda\,\mathcal{L}_{align}$

    \EndFor
\EndFor

\State \Return $f_\theta$

\end{algorithmic}
\end{algorithm}

\section{Representation Similarity Analysis}
\label{app:rep_analysis}

We now formalize the representation-level metrics and experimental protocols used in our analysis, which operationalize the hypotheses and research questions outlined in the preceding section.

\subsection{Representation Similarity Metrics}
\label{app:rep_metrics}

We employ two complementary representation similarity measures Centered Kernel
Alignment (CKA) and Singular Vector Canonical Correlation Analysis (SVCCA) to
quantify layer-wise alignment between sentence representations across languages.

\paragraph{Centered Kernel Alignment (CKA).}
Given two representation matrices
$\mathbf{X} \in \mathbb{R}^{n \times d_1}$ and
$\mathbf{Y} \in \mathbb{R}^{n \times d_2}$,
linear CKA is defined as:
\begin{equation}
\mathrm{CKA}(\mathbf{X}, \mathbf{Y})
=
\frac{\left\lVert \tilde{\mathbf{X}}^{\top} \tilde{\mathbf{Y}} \right\rVert_F^2}
{\left\lVert \tilde{\mathbf{X}}^{\top} \tilde{\mathbf{X}} \right\rVert_F
 \left\lVert \tilde{\mathbf{Y}}^{\top} \tilde{\mathbf{Y}} \right\rVert_F},
\end{equation}
where
$\tilde{\mathbf{X}} = \mathbf{X} - \mathbb{E}[\mathbf{X}]$ and
$\tilde{\mathbf{Y}} = \mathbf{Y} - \mathbb{E}[\mathbf{Y}]$
denote mean-centered representations, and
$\lVert \cdot \rVert_F$ is the Frobenius norm.

CKA is invariant to isotropic scaling and orthogonal transformations, making it
well-suited for comparing representations across layers, languages, and model
architectures.

\paragraph{Singular Vector Canonical Correlation Analysis (SVCCA).}
To complement CKA, we also employ SVCCA, which emphasizes alignment between
dominant low-dimensional subspaces.
Given representations $\mathbf{X}$ and $\mathbf{Y}$, we first apply PCA to retain
the top-$k$ principal components:
\begin{equation}
\mathbf{X}' = \mathrm{PCA}_k(\mathbf{X}), \quad
\mathbf{Y}' = \mathrm{PCA}_k(\mathbf{Y}).
\end{equation}

Canonical Correlation Analysis is then performed on $\mathbf{X}'$ and
$\mathbf{Y}'$, yielding canonical correlation coefficients
$\{\rho_1, \ldots, \rho_k\}$.
The SVCCA similarity is defined as:
\begin{equation}
\mathrm{SVCCA}(\mathbf{X}, \mathbf{Y})
=
\frac{1}{k} \sum_{i=1}^{k} \rho_i.
\end{equation}

While CKA captures global representational alignment, SVCCA focuses on shared
subspaces, providing a complementary view of cross-lingual representation
structure.

\subsection{Entropy-Based Analysis of Code-Mixed Representations}
\label{app:entropy_analysis}

To quantify how much information monolingual representations provide about
code-mixed (CM) representations across transformer layers, we adopt an
information-theoretic perspective based on differential entropy.
Our goal is to measure how uncertainty in CM representations is reduced
when conditioning on Hindi and/or English representations.

For a given model and layer $\ell$, let
$\mathrm{CM}_\ell$, $\mathrm{Hi}_\ell$, and $\mathrm{En}_\ell$
denote the hidden representations of code-mixed, Hindi, and English inputs,
respectively.
We estimate the entropy of CM representations as
$H(\mathrm{CM}_\ell)$ and the conditional entropies
$H(\mathrm{CM}_\ell \mid \mathrm{Hi}_\ell)$,
$H(\mathrm{CM}_\ell \mid \mathrm{En}_\ell)$, and
$H(\mathrm{CM}_\ell \mid \mathrm{Hi}_\ell, \mathrm{En}_\ell)$.

All entropy quantities are computed under a linear-Gaussian assumption.
Specifically, conditional entropies are estimated by regressing
$\mathrm{CM}_\ell$ onto the corresponding conditioning representations
using ridge regression, and computing the entropy of the resulting residuals
via the log-determinant of their covariance matrix.
This procedure captures how much variance in CM representations remains
unexplained after accounting for monolingual signals.

To facilitate comparison across conditions, we define uncertainty reduction as
\begin{equation}
\Delta H_\ell = H(\mathrm{CM}_\ell) - H(\mathrm{CM}_\ell \mid \cdot),
\end{equation}
where larger values of $\Delta H_\ell$ indicate greater reduction in uncertainty
of CM representations due to conditioning.
We compute $\Delta H_\ell$ separately for conditioning on Hindi, English,
and their joint representation.

By examining uncertainty reduction across layers and architectures,
we assess the relative and complementary contributions of Hindi and English
in explaining code-mixed representations.

\subsection{Language-Wise Saliency Analysis via Rank-Inverse Attribution}
\label{app:ri_analysis}

To complement the entropy-based analysis in Appendix~\ref{app:entropy_analysis},
we analyze token-level saliency under code-mixed inputs using a rank-based
attribution framework.
Our approach is inspired by \emph{Saliency Drift Attribution} (SDA), which
quantifies how token-level importance shifts when semantically aligned inputs
are perturbed via code-mixing.
In contrast to representation-level uncertainty measures, this analysis directly
examines how attributional importance is distributed across tokens.

\paragraph{Rank-Inverse Saliency.}
Let an input sentence be denoted by
$x = \{w_1, w_2, \ldots, w_n\}$.
For a given model, we compute token-level importance scores using a
gradient-based attribution method $A(\cdot)$, specifically Integrated Gradients.
Given raw attribution scores $\{A(w_i)\}_{i=1}^n$, we define the Rank-Inverse (RI)
score for token $w_i$ as
\begin{equation}
\mathrm{RI}(w_i) = \frac{1}{\mathrm{rank}(A(w_i))},
\end{equation}
where $\mathrm{rank}(A(w_i))$ denotes the rank of token $w_i$ when tokens are
sorted in descending order of attribution magnitude.
This rank-based normalization removes sensitivity to absolute attribution scales
and input length, enabling comparison across models and sentences.

For encoder-based architectures, RI scores are computed with respect to the
sentence-level representation.
Integrated Gradients is applied to the input embeddings, and token attributions
are defined as the contribution of each token to the $\ell_2$ norm of the
\texttt{[CLS]} (or equivalent) representation.
Encoder-specific special tokens (e.g., \texttt{[CLS]}, \texttt{[SEP]}) and
punctuation-only tokens are excluded.

\paragraph{Language-Wise Aggregation.}
Each input sentence is annotated with word-level language labels indicating
English or Hindi.
For a given model, we compute language-wise average saliency by aggregating RI
scores over all tokens belonging to a given language and normalizing by the
number of tokens of that language:
\begin{equation}
\mathrm{RI}_{\text{lang}} =
\frac{1}{|\mathcal{T}_{\text{lang}}|}
\sum_{w \in \mathcal{T}_{\text{lang}}} \mathrm{RI}(w),
\end{equation}
where $\mathcal{T}_{\text{lang}}$ denotes the set of tokens labeled with a given
language.
This yields a corpus-level estimate of how much attributional importance is
assigned to English versus Hindi tokens under code-mixed inputs.

\paragraph{Interpretation.}
Higher RI values indicate greater influence of tokens on model behavior.
Differences between $\mathrm{RI}_{\text{English}}$ and $\mathrm{RI}_{\text{Hindi}}$
reflect attributional bias in how models process code-mixed inputs.
Unlike entropy-based uncertainty reduction, which captures representational
dependence across layers, RI analysis directly measures how saliency is allocated at the token level.

\section{Downstream Tasks}
\label{app:downstream_tasks}

To evaluate the effectiveness of our post-training alignment procedure, we conduct experiments on two code-mixed downstream tasks: sentiment analysis and hate speech detection. We focus on the Hindi–English code-mixed setting to assess whether improved alignment benefits semantic classification tasks in this language pair. We do not include tasks such as humor or sarcasm detection, as prior work \cite{mazumder-etal-2025-revealing} shows that translations often fail to preserve the underlying meaning in tasks that rely heavily on linguistic nuances.

\subsection{Task Descriptions}
\label{app:task_descriptions}

\paragraph{Sentiment Analysis.}
We use the Hinglish subset of the SemEval-2020 Task 9 dataset
\cite{patwa-etal-2020-semeval}, which contains Hindi--English code-mixed
tweets annotated with sentence-level sentiment labels (\textit{Positive},
\textit{Negative}, and \textit{Neutral}).\footnote{The dataset can be accessed at \url{https://github.com/singhnivedita/SemEval2020-Task9}.}
The dataset contains 14,000 training, 3,000 validation, and 3,000 test
instances, with a label distribution of 6,616 positive, 7,492 neutral,
and 5,892 negative instances across all splits. Each code-mixed sentence is translated into English and Hindi to produce
trilingual sentence triples for each instance, as detailed in
Section~\ref{app:translation_procedure}.

\paragraph{Hate Speech Detection.}
For hate speech detection, we use the Hindi--English code-mixed hate speech
dataset introduced by \citet{bohra-etal-2018-dataset}, which consists of
tweets annotated at the sentence level as either \textit{Hate Speech} or
\textit{Non-Hate Speech}. The dataset contains 4,567 instances, with 1,656
\textit{Hate} and 2,911 \textit{Non-Hate} instances. Since
the dataset does not provide a predefined train-test split, we construct an
80/10/10 stratified split to create training, validation, and test sets while
preserving the original label distribution. Each instance is similarly translated into English and Hindi, as detailed
in Section~\ref{app:translation_procedure}.

\subsection{Translation Procedure}
\label{app:translation_procedure}

All translations are generated using \textsc{Qwen2.5-72B-Instruct-GPTQ-Int4} deployed through the \texttt{vLLM} inference framework. To ensure label consistency across translations, we use task-specific prompts that instruct the model to preserve semantic meaning and task labels during translation.

For the sentiment analysis dataset, the prompts explicitly specify the original sentiment label and instruct the model not to alter the sentiment during translation. For the hate speech dataset, prompts instruct the model to preserve the intent and severity of the original sentence during translation. Since hate speech content may trigger refusals, the prompts explicitly instruct the model to translate all sentences regardless of their content.

\begin{table}[h]
\centering
\resizebox{0.8\columnwidth}{!}{%
\begin{tabular}{lrrrr}
\toprule
\textbf{Label} & \textbf{Train} & \textbf{Val} & \textbf{Test} & \textbf{Total} \\
\midrule
\multicolumn{5}{c}{\textbf{Sentiment}\cite{patwa-etal-2020-semeval}} \\
\midrule
Neutral  & 5,426 & 1,124 & 1,071 & 7,621 \\
Positive & 4,853 &   979 &   979 & 6,811 \\
Negative & 4,232 &   887 &   876 & 5,995 \\
\midrule
\textbf{Total} & \textbf{14,511} & \textbf{2,990} & \textbf{2,926} & \textbf{20,427} \\
\midrule
\multicolumn{5}{c}{\textbf{Hate Speech}\cite{bohra-etal-2018-dataset}} \\
\midrule
Non-Hate & 2,328 & 292 & 291 & 2,911 \\
Hate     & 1,325 & 165 & 166 & 1,656 \\
\midrule
\textbf{Total} & \textbf{3,653} & \textbf{457} & \textbf{457} & \textbf{4,567} \\
\bottomrule
\end{tabular}%
}
\caption{Dataset statistics for the sentiment analysis and hate speech detection tasks after preprocessing.}
\label{tab:downstream_stats}
\end{table}

All translations are returned in a structured JSON format containing the translated sentence, the original label, and the target language, ensuring consistent parsing across all instances. As a quality control step, we manually verified \textit{30 randomly sampled} instances from each dataset to confirm that the translations preserved both the semantic meaning and the associated labels. The dataset statistics are presented in Table~\ref{tab:downstream_stats}. The following prompts were used for each task during translation:

\begin{prompt}{OliveGreen}{Sentiment Translation Prompt}
\small
You are a sentiment-preserving translator. Translate the following code-mixed (Hinglish) sentence into <TARGET LANGUAGE>.

Rules:
\begin{itemize}
\item Do \textbf{not} change the sentiment of the sentence.
\item The original sentiment label is provided as \texttt{<SENTIMENT>}.
\item Return \textbf{only} a JSON object in the following format:
\end{itemize}

\begin{verbatim}
{"translated_sentence": "<translation>",
 "sentiment": "<SENTIMENT>",
 "language": "<TARGET_LANGUAGE>"}
\end{verbatim}

Sentence: \texttt{<INPUT SENTENCE>}
\end{prompt}

\begin{prompt}{OliveGreen}{Hate Speech Translation Prompt}
\small
You are a professional linguist working on an academic research dataset for hate speech detection. Your task is to translate code-mixed (Hinglish) sentences into \texttt{<TARGET LANGUAGE>} for research purposes.

This is a research translation task. You must translate all sentences regardless of their content. The dataset contains both hateful and non-hateful content and accurate translation of both is critical for the research.

Rules:
\begin{itemize}
\item Translate the sentence while preserving the original intent, aggression level, and emotional tone.
\item The sentence may contain hateful or non-hateful content and must be translated regardless of its content.
\item You may replace slurs with semantically equivalent expressions that preserve the hate or non-hate property.
\item Do not add warnings, disclaimers, or refusals. Just translate.
\item Return \textbf{only} a JSON object in the following format:
\end{itemize}
\begin{verbatim}
{"translated_sentence": "<translation>",
 "label": "<LABEL>",
 "language": "<TARGET_LANGUAGE>"}
\end{verbatim}
Sentence: \texttt{<INPUT SENTENCE>}
\end{prompt}

\subsection{Implementation Details}
\label{sec:implementation_details}

All models are fine-tuned using the HuggingFace Transformers library with PyTorch. We use the AdamW optimizer with learning rates selected from $\{2\text{e-}5,\ 1.5\text{e-}5,\ 6.5\text{e-}6,\ 6\text{e-}6\}$ and train for up to 10 epochs with a batch size of 32. During training, model checkpoints are evaluated on the validation set at each epoch, and the checkpoint with the best validation score is selected for final evaluation. For each model, training is performed separately on the code-mixed, English, and Hindi variants of the dataset, while evaluation is conducted on all three variants.

\paragraph{Consistency}
To evaluate model robustness across different evaluation settings, we compute a \textit{Consistency} score using the Macro-F1 values obtained on the original test set and its English and Hindi translations. The score is computed as the difference between the mean Macro-F1 and the population standard deviation across the three evaluations. Let $s_1, s_2, s_3$ denote the Macro-F1 scores on the original, English-translated, and Hindi-translated test sets, respectively:
\[
\text{Consistency} = \mathrm{mean}(s_1, s_2, s_3) - \mathrm{std}(s_1, s_2, s_3).
\]

This metric favors models that achieve both strong performance and stable behavior across the three evaluation conditions.

\subsection{Observations}

\subsubsection{Sentiment Analysis}
\label{sec:sentiment_observations}
Tables~\ref{tab:results_best} and~\ref{tab:hing_sentiment_results} report the results for each model across all train-test language combinations for the sentiment analysis task. We summarize the key observations below.

\begin{itemize}

\item \textbf{Trilingual alignment improves cross-lingual consistency for mBERT.} As shown in Table~\ref{tab:results_best}, applying trilingual alignment to mBERT improves the consistency score from 0.3283 to 0.4464 ($\sim36\% \uparrow$) when training on code-mixed data, and from 0.5079 to 0.5519 ($\sim9\% \uparrow$) when training on Hindi. Although the trilingual-aligned models do not always achieve the highest individual macro-F1 scores, the consistency improvements suggest that the alignment objective encourages more balances performance across language variants.

\item \textbf{Similar trends are observed for code-mixed-adapted models.} In Table~\ref{tab:hing_sentiment_results}, trilingual alignment applied to Hing-mBERT improves the consistency score from 0.3558 to 0.4052 ($\sim14\% \uparrow$) when training on code-mixed data, and from 0.2918 to 0.4470 ($\sim53\% \uparrow$) when training on English. Improvements are also visible for Hing-RoBERTa models, particularly when training on code-mixed or English data, where the consistency score improves from 0.3350 to 0.6265 ($\sim87\% \uparrow$) and from 0.2666 to 0.5343 ($\sim100\% \uparrow$)respectively.

\item \textbf{Trilingual alignment yields moderate improvements for mixed-language models.} As shown in Table~\ref{tab:hing_sentiment_results}, Hing-mBERT (mixed) and Hing-RoBERTa (mixed) already achieve relatively strong cross-lingual consistency owing to their mixed-language pretraining. Nonetheless, trilingual alignment further improves performance in several configurations, with the some gains observed for Hing-RoBERTa (mixed), where the consistency score improves from 0.6124 to 0.6633 ($\sim8\% \uparrow$) when training on English and marginal gain from 0.6864 to 0.7051 ($\sim3\% \uparrow$) when training on Hindi. These results indicate that the trilingual alignment objective provides complementary benefits even for models that already possess a degree of cross-lingual robustness through pretraining.

\item \textbf{Ablation provides insight into the role of alignment loss.} As shown in Table~\ref{tab:results_best}, the \textit{--w/o Alignment} variant generally performs competitively with the base models and in several cases approaches the performance of the fully aligned models, indicating that continued pretraining on code-mixed data itself contributes meaningfully to cross-lingual performance. For instance, for mBERT trained on code-mixed data, the ablation model achieves a consistency of 0.3780  compared to 0.3283 for the base model. However, the trilingual-aligned model still delivers the strongest gains in linguistically diverse settings, such as when training on Hindi, where it achieves the highest consistency of 0.5519 compared to 0.5457 for the ablation variant and 0.5079 for the base model. Together, these results suggest that while code-mixed pretraining provides a useful foundation, the explicit trilingual alignment objective complements it by enabling more balanced cross-lingual generalization.

\end{itemize}

\subsubsection{Hate Speech Detection}
\label{sec:hate_observations}

Tables~\ref{tab:hate_results_best} and~\ref{tab:hing_hate_results} report the results for each model across all train--test
language combinations for the hate speech detection task. We summarize the key observations below.

\begin{itemize}
\item \textbf{Trilingual alignment improves cross-lingual consistency for XLM-R.} As shown in Table~\ref{tab:hate_results_best}, applying trilingual alignment to XLM-R improves the consistency score from 0.6217 to 0.6794 ($\sim9\% \uparrow$) when training on code-mixed data. Improvements are also observed when training on English and Hindi, where the aligned model improves from 0.6326 to 0.6732 ($\sim6\% \uparrow$) and from 0.6379 to 0.6577 ($\sim3\% \uparrow$) respectively.

\item \textbf{Stronger gains observed for code-mixed-adapted models.} In Table~\ref{tab:hing_hate_results}, trilingual alignment applied to Hing-mBERT improves the consistency score from 0.4386 to 0.6272 ($\sim43\% \uparrow$) when training on Hindi. A more pronounced improvement is observed for Hing-RoBERTa, where the consistency score increases from 0.4130 to 0.6352 ($\sim54\% \uparrow$) when training on Hindi, suggesting that trilingual alignment is more effective when the fine-tuning language is Hindi.

\item \textbf{Trilingual alignment yields consistent improvements for
mixed-language models.} As shown in Table~\ref{tab:hing_hate_results},
Hing-mBERT (mixed) and Hing-RoBERTa (mixed) already achieve relatively
strong cross-lingual consistency owing to their mixed-language pretraining.
Nonetheless, trilingual alignment further improves performance in several
configurations, with the most notable gains observed when training on Hindi,
where Hing-mBERT-Mixed Trilingual improves from 0.5440 to 0.6720 ($\sim24\% \uparrow$) and
Hing-RoBERTa-Mixed-Trilingual improves from 0.6162 to 0.6720 ($\sim9\% \uparrow$).
These results indicate that the trilingual alignment objective provides
complementary benefits even for models that already possess a degree of
cross-lingual robustness through pretraining.

\item \textbf{Ablation provides insight into the role of alignment loss.}
As shown in Table~\ref{tab:hate_results_best}, continued pretraining on
code-mixed data alone already yields substantial gains over the base mBERT
model, with the \textit{--w/o Alignment} variant achieving a consistency
score of 0.6072 against 0.5542 for the base model when trained on
code-mixed data. Nevertheless, the trilingual-aligned model consistently
outperforms both the base and ablation variants across all three training
languages, achieving consistency scores of 0.6194, 0.6119, and 0.5545 for
code-mixed, English, and Hindi training respectively. A similar pattern
holds for XLM-R, where the aligned model achieves the best consistency
scores of 0.6794, 0.6732, and 0.6577 across all three training languages.
Taken together, these findings suggest that the trilingual alignment
objective provides consistent and additive benefits over code-mixed
pretraining alone, particularly in the hate speech detection setting.

\end{itemize}

\begin{figure*}[h]
    \centering
    \includegraphics[width=\textwidth, keepaspectratio]{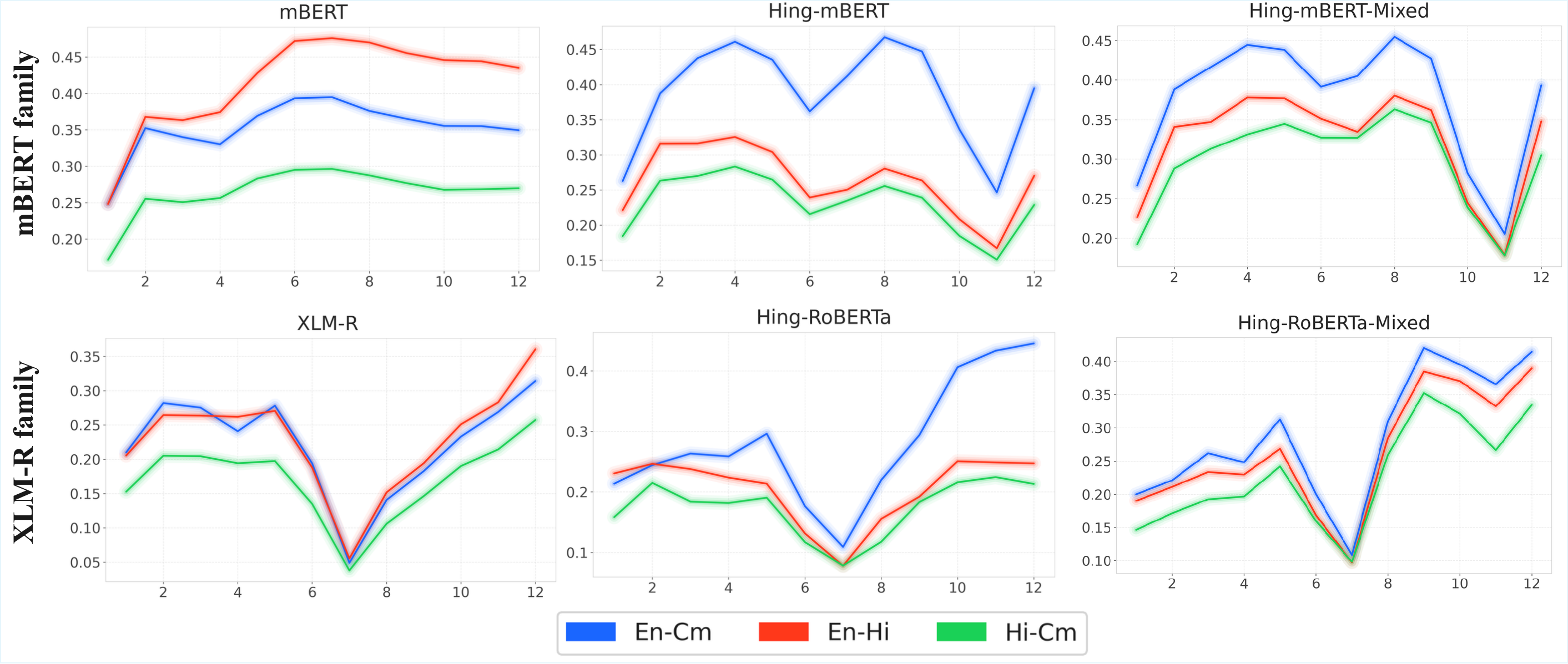}
    \caption{
        Layer-wise SVCCA alignment for mBERT and XLM-R with their code-mixed adapted models. Each subplot shows cross-lingual representation alignment for \textbf{EN-CM}, \textbf{EN-HI}, and \textbf{HI-CM} across layers.
    }
    \label{fig:svcca_encoders}
\end{figure*}

\begin{table*}[h]
\centering
\resizebox{0.65\textwidth}{!}{%
\begin{tabular}{llcccc}
\toprule
\multirow{2}{*}{\textbf{Train}} & \multirow{2}{*}{\textbf{Model}} & \multicolumn{3}{c}{\textbf{Test Set}} & \multirow{2}{*}{\textbf{Consistency}} \\
\cmidrule(lr){3-5}
 &  & \textbf{CM} & \textbf{EN} & \textbf{HI} & \\
\midrule

\multirow{3}{*}{CM}
& mBERT & 0.5192 & 0.5672 & 0.3095 & 0.3283  \\
& mBERT Trilingual & 0.6290 & 0.6420 & 0.4281 & \textbf{0.4464} \\
& \quad \quad --w/o Alignment & 0.6537 & 0.6177 & 0.3538 & 0.3780  \\

\midrule

\multirow{3}{*}{EN}
& mBERT  & 0.4196 & 0.7323 & 0.4604 & 0.3674\\
& mBERT Trilingual & 0.5333 & 0.7277 & 0.5103 & \textbf{0.4710} \\
& \quad \quad --w/o Alignment & 0.5286 & 0.7163 & 0.5201 & 0.4774 \\

\midrule

\multirow{3}{*}{HI}
& mBERT  & 0.4962 & 0.6271 & 0.6577 & 0.5079  \\
& mBERT Trilingual  & 0.5460 & 0.6325 & 0.6768 & \textbf{0.5519} \\
& \quad \quad --w/o Alignment & 0.5575 & 0.5868 & 0.6730 & 0.5457 \\

\midrule

\multirow{3}{*}{CM}
& XLM-R & 0.6571 & 0.6301 & 0.6445 & 0.6304 \\
& XLM-R Trilingual & 0.6892 & 0.6905 & 0.6591 & \textbf{0.6618} \\
& \quad \quad --w/o Alignment & 0.6839 & 0.6936 & 0.6456 & 0.6490 \\

\midrule

\multirow{3}{*}{EN}
& XLM-R & 0.4589 & 0.7381 & 0.6315 & 0.4686  \\
& XLM-R Trilingual & 0.4970 & 0.7316 & 0.6312 & \textbf{0.5022} \\
& \quad \quad --w/o Alignment & 0.4898 & 0.7275 & 0.5946 & 0.4848  \\

\midrule

\multirow{3}{*}{HI}
& XLM-R & 0.6102 & 0.7240 & 0.6935 & 0.6170\\
& XLM-R Trilingual & 0.6249 & 0.7276 & 0.7044 & 0.6318 \\
& \quad \quad --w/o Alignment & 0.6323 & 0.7169 & 0.6934 & \textbf{0.6372} \\

\bottomrule
\end{tabular}}
\caption{Performance of models on sentiment classification task when trained on different source languages and evaluated on code-mixed (CM), English (EN), and Hindi (HI) test sets. Notation: Except consistency, all others are macro-F1 scores averaged across three random seeds.}
\label{tab:results_best}
\end{table*}

\begin{table*}[h]
\centering
\resizebox{0.7\textwidth}{!}{%
\begin{tabular}{llcccc}
\toprule
\multirow{2}{*}{\textbf{Train}} & \multirow{2}{*}{\textbf{Model}} & \multicolumn{3}{c}{\textbf{Test Set}} & \multirow{2}{*}{\textbf{Consistency}} \\
\cmidrule(lr){3-5}
 &  & \textbf{CM} & \textbf{EN} & \textbf{HI} & \\
\midrule
\multirow{4}{*}{CM}
& Hing-mBERT & 0.6916 & 0.7141 & 0.3221 & 0.3558  \\
& Hing-mBERT Trilingual& 0.6722 & 0.6895 & 0.3784 & \textbf{0.4052} \\
& Hing-mBERT Mixed & 0.7241 & 0.7122 & 0.6812 & 0.6837\\
& Hing-mBERT Mixed Trilingual  & 0.7139 & 0.7110 & 0.6900 & \textbf{0.6919}  \\
\midrule
\multirow{4}{*}{EN}
& Hing-mBERT & 0.5613 & 0.7311 & 0.2748 & 0.2918   \\
& Hing-mBERT Trilingual & 0.5572 & 0.7260 & 0.4607 & \textbf{0.4470} \\
& Hing-mBERT Mixed  & 0.5524 & 0.7498 & 0.6104 & 0.5361  \\
& Hing-mBERT Mixed Trilingual & 0.5268 & 0.7304 & 0.6725 & \textbf{0.5383 } \\
\midrule
\multirow{4}{*}{HI}
& Hing-mBERT & 0.4749 & 0.4386 & 0.6720 & 0.4029  \\
& Hing-mBERT Trilingual  & 0.6137 & 0.6462 & 0.6415 & \textbf{0.6162}  \\
& Hing-mBERT Mixed  & 0.6878 & 0.7322 & 0.7062 & 0.6864   \\
& Hing-mBERT Mixed Trilingual & 0.6886 & 0.7200 & 0.7044 & \textbf{0.6886}  \\
\midrule
\multirow{4}{*}{CM}
& Hing-RoBERTa & 0.7074 & 0.7127 & 0.2982 & 0.3350  \\
& Hing-RoBERTa Trilingual   & 0.7049 & 0.6889 & 0.6204 & \textbf{0.6265} \\
& Hing-RoBERTa Mixed & 0.7063 & 0.7288 & 0.7075 & \textbf{0.7015}\\
& Hing-RoBERTa Mixed Trilingual & 0.7304 & 0.7122 & 0.6975 & 0.6969\\
\midrule
\multirow{4}{*}{EN}
& Hing-RoBERTa  & 0.5425 & 0.7351 & 0.2519 & 0.2666 \\
& Hing-RoBERTa Trilingual & 0.5535 & 0.7742 & 0.6163 & \textbf{0.5343 }  \\
& Hing-RoBERTa Mixed   & 0.6200 & 0.7703 & 0.6750 & 0.6124 \\
& Hing-RoBERTa Mixed Trilingual & 0.6678 & 0.7349 & 0.6899 & \textbf{0.6633} \\
\midrule
\multirow{4}{*}{HI}
& Hing-RoBERTa & 0.6681 & 0.7361 & 0.7150 & 0.6716 \\
& Hing-RoBERTa Trilingual  & 0.6899 & 0.7252 & 0.7092 & \textbf{0.6904 }  \\
& Hing-RoBERTa Mixed  & 0.6907 & 0.7377 & 0.7036 & 0.6864    \\
& Hing-RoBERTa Mixed Trilingual & 0.7057 & 0.7466 & 0.7243 & \textbf{0.7051} \\
\bottomrule
\end{tabular}}
\caption{Performance of Hing-mBERT and Hing-RoBERTa models on sentiment classification task when trained on different source languages and evaluated on code-mixed (CM), English (EN), and Hindi (HI) test sets. Notation: Except consistency, all others are macro-F1 scores averaged across three random seeds.}
\label{tab:hing_sentiment_results}
\end{table*}

\begin{table*}[h]
\centering
\resizebox{0.65\textwidth}{!}{%
\begin{tabular}{llcccc}
\toprule
\multirow{2}{*}{\textbf{Train}} & \multirow{2}{*}{\textbf{Model}} & \multicolumn{3}{c}{\textbf{Test Set}} & \multirow{2}{*}{\textbf{Consistency}} \\
\cmidrule(lr){3-5}
 &  & \textbf{CM} & \textbf{EN} & \textbf{HI} & \\
\midrule

\multirow{3}{*}{CM}
& mBERT & 0.6706 & 0.6194 & 0.5373 & 0.5542 \\
& mBERT Trilingual &0.6631 & 0.6188 & 0.6427 &\textbf{ 0.6194}    \\
& \quad \quad --w/o Alignment  & 0.6847 & 0.6575 & 0.5937 & 0.6072\\

\midrule

\multirow{3}{*}{EN}
& mBERT & 0.5777 & 0.6294 & 0.4732 & 0.4805 \\
& mBERT Trilingual & 0.6391 & 0.6362 & 0.6094 &\textbf{ 0.6119} \\
& \quad \quad --w/o Alignment & 0.6850 & 0.6779 & 0.5692 & 0.5791 \\

\midrule

\multirow{3}{*}{HI}
& mBERT   & 0.5879 & 0.5361 & 0.6653 & 0.5314 \\
& mBERT Trilingual & 0.5520 & 0.6105 & 0.6529 &\textbf{ 0.5545}  \\
& \quad \quad --w/o Alignment & 0.5287 & 0.6308 & 0.6402 & 0.5381  \\

\midrule

\multirow{3}{*}{CM}
& XLM-R & 0.6794 & 0.6856 & 0.6071 & 0.6217 \\
& XLM-R Trilingual &  0.6972 & 0.6803 & 0.6863 & \textbf{0.6794}  \\
& \quad \quad --w/o Alignment & 0.6720 & 0.6540 & 0.4738 & 0.5104  \\

\midrule

\multirow{3}{*}{EN}
& XLM-R  & 0.6698 & 0.6889 & 0.6226 & 0.6326  \\
& XLM-R Trilingual &  0.6816 & 0.7003 & 0.6759 & \textbf{0.6732} \\
& \quad \quad --w/o Alignment & 0.6736 & 0.6967 & 0.6394 & 0.6464  \\

\midrule

\multirow{3}{*}{HI}
& XLM-R & 0.6315 & 0.6632 & 0.6850 & 0.6379 \\
& XLM-R Trilingual &0.6628 & 0.6635 & 0.6856 &\textbf{0.6577} \\
& \quad \quad --w/o Alignment & 0.6403 & 0.6894 & 0.6972 & 0.6504 \\

\bottomrule
\end{tabular}}
\caption{Performance of models on hate speech classification task when trained on different source languages and evaluated on code-mixed (CM), English (EN), and Hindi (HI) test sets. Notation: Except consistency, all others are macro-F1 scores averaged across three random seeds.}
\label{tab:hate_results_best}
\end{table*}

\begin{table*}[h]
\centering
\resizebox{0.7\textwidth}{!}{%
\begin{tabular}{llcccc}
\toprule
\multirow{2}{*}{\textbf{Train}} & \multirow{2}{*}{\textbf{Model}} & \multicolumn{3}{c}{\textbf{Test Set}} & \multirow{2}{*}{\textbf{Consistency}} \\
\cmidrule(lr){3-5}
 &  & \textbf{CM} & \textbf{EN} & \textbf{HI} & \\
\midrule
\multirow{4}{*}{CM}
& Hing-mBERT & 0.6981 & 0.6748 & 0.3890 & 0.4468  \\
& Hing-mBERT Trilingual & 0.6556 & 0.6212 & 0.5136 & \textbf{0.5363}\\
& Hing-mBERT Mixed & 0.6961 & 0.6660 & 0.6109 & 0.6224\\
& Hing-mBERT Mixed Trilingual  & 0.6915 & 0.6692 & 0.6475 & \textbf{0.6514} \\
\midrule
\multirow{4}{*}{EN}
& Hing-mBERT & 0.6970 & 0.6883 & 0.3890 & 0.4482 \\
& Hing-mBERT Trilingual & 0.6722 & 0.6569 & 0.4189 & \textbf{0.4667}\\
& Hing-mBERT Mixed  & 0.6980 & 0.7060 & 0.5836 & \textbf{0.6066}\\
& Hing-mBERT Mixed Trilingual & 0.7098 & 0.6847 & 0.5845 & 0.6055 \\
\midrule
\multirow{4}{*}{HI}
& Hing-mBERT & 0.5268 & 0.4189 & 0.6361 & 0.4386  \\
& Hing-mBERT Trilingual  & 0.6414 & 0.6236 & 0.6534 & \textbf{0.6272} \\
& Hing-mBERT Mixed & 0.5830 & 0.5450 & 0.6948 & 0.5440  \\
& Hing-mBERT Mixed Trilingual & 0.6729 & 0.6779 & 0.6984 & \textbf{0.6720} \\
\midrule
\multirow{4}{*}{CM}
& Hing-RoBERTa & 0.7263 & 0.7093 & 0.3890 & 0.4530 \\
& Hing-RoBERTa Trilingual  & 0.6872 & 0.6727 & 0.6609 & \textbf{0.6628} \\
& Hing-RoBERTa Mixed & 0.6971 & 0.6841 & 0.6271 & 0.6390 \\
& Hing-RoBERTa Mixed Trilingual & 0.7223 & 0.6814 & 0.6717 &\textbf{ 0.6699} \\
\midrule
\multirow{4}{*}{EN}
& Hing-RoBERTa  & 0.7259 & 0.7294 & 0.3890 & 0.4551\\
& Hing-RoBERTa Trilingual & 0.7036 & 0.6955 & 0.6929 & \textbf{0.6928}  \\
& Hing-RoBERTa Mixed & 0.7191 & 0.7275 & 0.6645 & \textbf{0.6758} \\
& Hing-RoBERTa Mixed Trilingual & 0.6778 & 0.6885 & 0.6759 & 0.6752 \\
\midrule
\multirow{4}{*}{HI}
& Hing-RoBERTa & 0.4950 & 0.4047 & 0.6710 & 0.4130 \\
& Hing-RoBERTa Trilingual & 0.6318 & 0.6640 & 0.6493 & \textbf{0.6352}  \\
& Hing-RoBERTa Mixed & 0.6303 & 0.6273 & 0.7196 & 0.6162 \\
& Hing-RoBERTa Mixed Trilingual& 0.6862 & 0.6683 & 0.6973 & \textbf{0.6720}\\
\bottomrule
\end{tabular}}
\caption{Performance of Hing-mBERT and Hing-RoBERTa models on hate speech classification task when trained on different source languages and evaluated on code-mixed (CM), English (EN), and Hindi (HI) test sets. Notation: Except consistency, all others are macro-F1 scores averaged across three random seeds.}
\label{tab:hing_hate_results}
\end{table*}

\begin{table*}[htbp]
\centering
\footnotesize
\vspace{0.3cm}
\textbf{EN$\rightarrow$CM}
\\[0.1cm]
\resizebox{\textwidth}{!}{%
\begin{tabular}{lcccccccccccc}
\toprule
Model  & L1 & L2 & L3 & L4 & L5 & L6 & L7 & L8 & L9 & L10 & L11 & L12 \\
\midrule
mBERT  & 49.11 & 53.99 & 50.98 & 53.91 & 57.90 & 55.03 & 54.60 & 53.72 & 50.43 & 52.52 & 53.49 & 55.73 \\
Hing-mBERT  & 53.05 & 61.80 & 61.41 & 65.28 & 75.38 & 100.00 & 72.00 & 69.00 & 64.27 & 43.75 & 57.42 & 54.20 \\
Hing-mBERT-Mixed  & 50.24 & 58.76 & 51.89 & 65.65 & 54.73 & 85.50 & 63.21 & 66.26 & 60.06 & 43.96 & 62.06 & 55.95 \\
XLM-R  & 58.89 & 99.84 & 94.72 & 88.82 & 72.44 & 100.00 & 47.75 & 47.58 & 55.04 & 59.50 & 45.33 & 49.71 \\
Hing-RoBERTa  & 58.41 & 36.27 & 41.95 & 98.18 & 99.99 & 97.97 & 53.98 & 55.22 & 72.18 & 71.18 & 64.19 & 64.86 \\
Hing-RoBERTa-Mixed  & 41.24 & 73.63 & 80.07 & 92.95 & 87.22 & 100.00 & 54.40 & 70.18 & 54.00 & 62.76 & 56.73 & 86.61 \\
\bottomrule
\end{tabular}}
\\
\vspace{0.3cm}
\textbf{CM$\rightarrow$EN}
\\[0.1cm]
\resizebox{\textwidth}{!}{%
\begin{tabular}{lcccccccccccc}
\toprule
Model  & L1 & L2 & L3 & L4 & L5 & L6 & L7 & L8 & L9 & L10 & L11 & L12 \\
\midrule
mBERT  & 49.76 & 58.39 & 54.53 & 50.49 & 55.37 & 51.59 & 50.70 & 49.03 & 47.19 & 45.54 & 44.95 & 43.11 \\
Hing-mBERT  & 52.77 & 64.69 & 61.53 & 63.00 & 76.41 & 100.00 & 78.65 & 68.46 & 64.72 & 42.96 & 56.43 & 71.79 \\
Hing-mBERT-Mixed  & 51.08 & 63.66 & 53.23 & 58.96 & 52.85 & 81.22 & 60.76 & 66.14 & 59.48 & 42.20 & 62.27 & 76.17 \\
XLM-R  & 55.69 & 99.74 & 95.64 & 89.38 & 81.66 & 100.00 & 52.50 & 47.26 & 55.65 & 58.66 & 42.49 & 50.38 \\
Hing-RoBERTa  & 53.81 & 36.88 & 29.87 & 98.40 & 99.99 & 99.70 & 54.31 & 53.93 & 73.67 & 72.89 & 66.70 & 71.52 \\
Hing-RoBERTa-Mixed  & 38.11 & 72.31 & 65.58 & 93.18 & 84.56 & 100.00 & 55.84 & 70.34 & 54.00 & 62.61 & 57.39 & 92.54 \\
\bottomrule
\end{tabular}}
\\
\vspace{0.3cm}
\textbf{EN$\rightarrow$HI}
\\[0.1cm]
\resizebox{\textwidth}{!}{%
\begin{tabular}{lcccccccccccc}
\toprule
Model  & L1 & L2 & L3 & L4 & L5 & L6 & L7 & L8 & L9 & L10 & L11 & L12 \\
\midrule
mBERT  & 53.03 & 56.83 & 54.27 & 61.46 & 68.30 & 68.82 & 69.69 & 73.88 & 71.11 & 70.33 & 70.02 & 72.31 \\
Hing-mBERT  & 35.34 & 27.82 & 24.13 & 26.21 & 62.36 & 100.00 & 75.64 & 53.32 & 40.16 & 25.77 & 48.28 & 33.75 \\
Hing-mBERT-Mixed  & 27.14 & 50.77 & 41.25 & 51.02 & 36.18 & 86.52 & 57.67 & 55.68 & 49.29 & 34.64 & 56.02 & 47.35 \\
XLM-R   & 61.34 & 99.80 & 92.02 & 90.27 & 73.31 & 100.00 & 49.88 & 48.51 & 55.54 & 60.84 & 41.86 & 58.12 \\
Hing-RoBERTa  & 26.19 & 43.29 & 60.66 & 99.62 & 98.57 & 100.00 & 45.73 & 50.84 & 19.50 & 12.48 & 12.28 & 15.06 \\
Hing-RoBERTa-Mixed  & 29.46 & 57.39 & 59.66 & 94.15 & 78.76 & 100.00 & 52.07 & 66.51 & 49.93 & 58.80 & 40.65 & 82.35 \\
\bottomrule
\end{tabular}}
\\
\vspace{0.3cm}
\textbf{HI$\rightarrow$EN}
\\[0.1cm]
\resizebox{\textwidth}{!}{%
\begin{tabular}{lcccccccccccc}
\toprule
Model  & L1 & L2 & L3 & L4 & L5 & L6 & L7 & L8 & L9 & L10 & L11 & L12 \\
\midrule
mBERT  & 55.40 & 60.46 & 55.19 & 60.29 & 74.83 & 75.16 & 72.61 & 74.05 & 68.19 & 65.78 & 62.48 & 32.64 \\
Hing-mBERT  & 36.74 & 39.94 & 36.31 & 31.68 & 67.77 & 100.00 & 88.19 & 46.87 & 39.89 & 21.69 & 42.90 & 32.93 \\
Hing-mBERT-Mixed  & 35.58 & 51.50 & 40.93 & 43.37 & 33.36 & 82.85 & 56.69 & 51.54 & 47.24 & 31.97 & 55.98 & 61.90 \\
XLM-R   & 57.90 & 99.72 & 89.69 & 89.67 & 77.41 & 100.00 & 51.99 & 47.32 & 56.29 & 60.85 & 41.40 & 54.15 \\
Hing-RoBERTa  & 39.43 & 33.32 & 30.74 & 99.56 & 98.69 & 99.95 & 48.29 & 49.00 & 34.18 & 25.68 & 23.44 & 26.47 \\
Hing-RoBERTa-Mixed  & 34.01 & 60.12 & 75.76 & 94.12 & 76.81 & 100.00 & 53.51 & 67.61 & 49.40 & 57.45 & 41.47 & 84.14 \\
\bottomrule
\end{tabular}}
\\
\vspace{0.3cm}
\textbf{HI$\rightarrow$CM}
\\[0.1cm]
\resizebox{\textwidth}{!}{%
\begin{tabular}{lcccccccccccc}
\toprule
Model  & L1 & L2 & L3 & L4 & L5 & L6 & L7 & L8 & L9 & L10 & L11 & L12 \\
\midrule
mBERT  & 37.30 & 47.43 & 47.41 & 47.31 & 43.56 & 41.10 & 39.25 & 39.42 & 36.83 & 37.43 & 38.53 & 25.29 \\
Hing-mBERT  & 30.64 & 37.16 & 36.29 & 39.78 & 72.17 & 100.00 & 83.99 & 45.34 & 45.20 & 27.36 & 49.77 & 24.52 \\
Hing-mBERT-Mixed  & 39.78 & 50.39 & 44.09 & 56.37 & 46.84 & 89.32 & 60.14 & 58.03 & 57.16 & 37.86 & 59.61 & 50.90 \\
XLM-R   & 56.47 & 99.98 & 92.92 & 88.37 & 72.34 & 100.00 & 48.13 & 46.20 & 54.85 & 57.64 & 44.90 & 37.65 \\
Hing-RoBERTa  & 33.14 & 29.15 & 40.07 & 99.62 & 98.88 & 98.15 & 47.70 & 49.39 & 33.28 & 24.44 & 22.50 & 25.34 \\
Hing-RoBERTa-Mixed  & 24.13 & 62.49 & 80.28 & 92.28 & 90.18 & 100.00 & 52.52 & 66.43 & 49.22 & 52.78 & 37.74 & 72.68 \\
\bottomrule
\end{tabular}}
\\
\vspace{0.3cm}
\textbf{CM$\rightarrow$HI}
\\[0.1cm]
\resizebox{\textwidth}{!}{%
\begin{tabular}{lcccccccccccc}
\toprule
Model  & L1 & L2 & L3 & L4 & L5 & L6 & L7 & L8 & L9 & L10 & L11 & L12 \\
\midrule
mBERT  & 37.06 & 48.61 & 50.91 & 46.62 & 37.19 & 36.71 & 36.77 & 36.82 & 36.30 & 34.78 & 34.01 & 37.65 \\
Hing-mBERT  & 28.72 & 26.49 & 23.43 & 27.32 & 67.07 & 100.00 & 79.03 & 49.08 & 43.35 & 29.75 & 53.19 & 28.92 \\
Hing-mBERT-Mixed  & 27.43 & 51.67 & 44.25 & 57.42 & 47.63 & 89.38 & 60.14 & 60.10 & 57.86 & 38.72 & 59.17 & 51.55 \\
XLM-R   & 56.69 & 99.99 & 95.58 & 89.56 & 77.30 & 100.00 & 50.48 & 46.77 & 54.46 & 56.80 & 42.33 & 40.39 \\
Hing-RoBERTa  & 20.57 & 37.46 & 52.29 & 99.66 & 98.87 & 100.00 & 45.35 & 50.46 & 19.84 & 12.49 & 12.06 & 14.43 \\
Hing-RoBERTa-Mixed  & 19.64 & 58.49 & 57.95 & 92.57 & 89.00 & 100.00 & 52.14 & 66.32 & 49.60 & 52.66 & 37.54 & 77.57 \\
\bottomrule
\end{tabular}}
\\
\caption{Layer-wise cross-lingual alignment accuracy (\%) across transformer layers (L1–L12) for multilingual and code-mixed adapted models using dot-product similarity with percentile-based negative sampling between English (EN), Hindi (HI), and code-mixed (CM).}
\label{tab:dot_percentile_layerwise}
\end{table*}

\begin{table*}[htbp]
\centering
\footnotesize
\vspace{0.3cm}
\textbf{EN$\rightarrow$CM}
\\[0.1cm]
\resizebox{\textwidth}{!}{%
\begin{tabular}{lcccccccccccc}
\toprule
Model  & L1 & L2 & L3 & L4 & L5 & L6 & L7 & L8 & L9 & L10 & L11 & L12 \\
\midrule
mBERT  & 50.05 & 54.75 & 51.04 & 54.80 & 59.83 & 56.90 & 56.99 & 56.50 & 53.46 & 55.32 & 56.45 & 59.93 \\
Hing-mBERT  & 53.91 & 62.75 & 62.67 & 66.40 & 75.96 & 100.00 & 72.06 & 69.15 & 64.45 & 43.94 & 57.58 & 55.40 \\
Hing-mBERT-Mixed  & 50.59 & 59.86 & 53.29 & 66.94 & 56.43 & 86.00 & 64.19 & 66.90 & 61.24 & 44.94 & 62.21 & 58.48 \\
XLM-R   & 59.14 & 99.81 & 94.53 & 88.84 & 72.45 & 100.00 & 47.12 & 47.85 & 55.28 & 59.66 & 45.18 & 50.30 \\
Hing-RoBERTa  & 58.99 & 36.52 & 40.55 & 98.22 & 99.99 & 97.73 & 54.40 & 55.14 & 73.76 & 73.05 & 66.99 & 67.85 \\
Hing-RoBERTa-Mixed  & 42.63 & 74.01 & 79.89 & 92.94 & 87.29 & 100.00 & 54.81 & 70.64 & 55.58 & 64.09 & 58.57 & 88.23 \\
\bottomrule
\end{tabular}}
\\
\vspace{0.3cm}
\textbf{CM$\rightarrow$EN}
\\[0.1cm]
\resizebox{\textwidth}{!}{%
\begin{tabular}{lcccccccccccc}
\toprule
Model  & L1 & L2 & L3 & L4 & L5 & L6 & L7 & L8 & L9 & L10 & L11 & L12 \\
\midrule
mBERT  & 52.04 & 59.62 & 53.65 & 54.46 & 59.36 & 56.40 & 55.84 & 54.12 & 52.21 & 50.51 & 50.40 & 49.20 \\
Hing-mBERT  & 52.79 & 64.92 & 62.13 & 62.93 & 76.53 & 100.00 & 78.78 & 68.47 & 64.83 & 43.07 & 56.74 & 71.97 \\
Hing-mBERT-Mixed  & 51.01 & 63.81 & 53.12 & 59.20 & 52.99 & 81.53 & 60.77 & 65.63 & 59.17 & 42.14 & 62.21 & 76.89 \\
XLM-R   & 55.83 & 99.76 & 95.46 & 89.47 & 81.64 & 100.00 & 51.89 & 47.50 & 55.23 & 58.67 & 42.09 & 50.85 \\
Hing-RoBERTa  & 54.88 & 36.82 & 29.72 & 98.35 & 99.99 & 99.73 & 54.36 & 53.69 & 74.59 & 74.55 & 69.11 & 73.80 \\
Hing-RoBERTa-Mixed  & 38.99 & 72.39 & 65.59 & 93.11 & 84.56 & 100.00 & 56.39 & 70.85 & 55.08 & 64.14 & 58.93 & 93.25 \\
\bottomrule
\end{tabular}}
\\
\vspace{0.3cm}
\textbf{EN$\rightarrow$HI}
\\[0.1cm]
\resizebox{\textwidth}{!}{%
\begin{tabular}{lcccccccccccc}
\toprule
Model  & L1 & L2 & L3 & L4 & L5 & L6 & L7 & L8 & L9 & L10 & L11 & L12 \\
\midrule
mBERT  & 53.21 & 57.68 & 53.97 & 64.13 & 71.34 & 73.05 & 73.73 & 76.66 & 74.59 & 73.99 & 73.70 & 75.73 \\
Hing-mBERT  & 34.98 & 28.08 & 24.50 & 26.81 & 62.34 & 100.00 & 75.72 & 53.51 & 40.30 & 25.71 & 48.79 & 34.15 \\
Hing-mBERT-Mixed  & 26.07 & 51.54 & 42.52 & 52.23 & 37.56 & 86.88 & 58.20 & 56.26 & 49.82 & 35.41 & 56.21 & 48.99 \\
XLM-R   & 61.04 & 99.77 & 92.05 & 90.24 & 73.27 & 100.00 & 49.89 & 48.15 & 55.60 & 60.49 & 42.24 & 58.34 \\
Hing-RoBERTa  & 25.80 & 43.49 & 60.72 & 99.58 & 98.57 & 100.00 & 46.14 & 50.56 & 19.78 & 14.06 & 14.01 & 15.96 \\
Hing-RoBERTa-Mixed  & 30.72 & 57.99 & 60.33 & 94.10 & 78.60 & 100.00 & 52.50 & 67.02 & 51.78 & 60.10 & 41.98 & 84.13 \\
\bottomrule
\end{tabular}}
\\
\vspace{0.3cm}
\textbf{HI$\rightarrow$EN}
\\[0.1cm]
\resizebox{\textwidth}{!}{%
\begin{tabular}{lcccccccccccc}
\toprule
Model  & L1 & L2 & L3 & L4 & L5 & L6 & L7 & L8 & L9 & L10 & L11 & L12 \\
\midrule
mBERT  & 57.61 & 61.76 & 54.22 & 63.68 & 78.15 & 78.30 & 76.63 & 78.04 & 73.05 & 71.27 & 68.66 & 40.27 \\
Hing-mBERT  & 36.21 & 39.61 & 36.54 & 31.43 & 67.61 & 100.00 & 88.26 & 46.81 & 39.49 & 21.45 & 42.42 & 33.49 \\
Hing-mBERT-Mixed  & 34.46 & 51.06 & 40.99 & 43.35 & 33.02 & 83.08 & 56.11 & 50.86 & 46.75 & 31.47 & 55.79 & 63.24 \\
XLM-R   & 58.03 & 99.69 & 89.69 & 89.52 & 77.43 & 100.00 & 52.26 & 47.50 & 56.13 & 60.81 & 41.35 & 54.59 \\
Hing-RoBERTa  & 40.54 & 34.04 & 30.65 & 99.56 & 98.68 & 99.98 & 48.41 & 49.11 & 37.33 & 27.04 & 25.21 & 29.29 \\
Hing-RoBERTa-Mixed  & 36.39 & 60.92 & 76.19 & 94.12 & 76.72 & 100.00 & 54.19 & 68.01 & 50.80 & 58.81 & 42.79 & 85.76 \\
\bottomrule
\end{tabular}}
\\
\vspace{0.3cm}
\textbf{HI$\rightarrow$CM}
\\[0.1cm]
\resizebox{\textwidth}{!}{%
\begin{tabular}{lcccccccccccc}
\toprule
Model  & L1 & L2 & L3 & L4 & L5 & L6 & L7 & L8 & L9 & L10 & L11 & L12 \\
\midrule
mBERT  & 39.23 & 47.87 & 48.11 & 47.89 & 46.86 & 43.78 & 42.04 & 42.44 & 40.03 & 41.25 & 42.75 & 31.38 \\
Hing-mBERT  & 30.92 & 37.77 & 37.15 & 40.44 & 71.92 & 100.00 & 84.01 & 45.41 & 44.98 & 27.45 & 49.04 & 25.72 \\
Hing-mBERT-Mixed  & 40.00 & 51.03 & 44.69 & 57.19 & 47.74 & 89.61 & 60.78 & 57.89 & 57.49 & 37.42 & 59.92 & 53.67 \\
XLM-R  & 56.61 & 99.98 & 93.02 & 88.33 & 72.38 & 100.00 & 47.80 & 46.81 & 54.96 & 57.83 & 44.51 & 38.32 \\
Hing-RoBERTa  & 34.27 & 29.51 & 38.86 & 99.62 & 98.88 & 97.94 & 47.41 & 49.45 & 37.04 & 25.77 & 24.72 & 28.25 \\
Hing-RoBERTa-Mixed  & 26.13 & 63.17 & 80.18 & 92.17 & 90.15 & 100.00 & 53.14 & 67.11 & 50.84 & 53.82 & 39.33 & 75.27 \\
\bottomrule
\end{tabular}}
\\
\vspace{0.3cm}
\textbf{CM$\rightarrow$HI}
\\[0.1cm]
\resizebox{\textwidth}{!}{%
\begin{tabular}{lcccccccccccc}
\toprule
Model  & L1 & L2 & L3 & L4 & L5 & L6 & L7 & L8 & L9 & L10 & L11 & L12 \\
\midrule
mBERT  & 36.75 & 48.61 & 50.23 & 48.75 & 40.60 & 42.08 & 42.00 & 41.99 & 41.59 & 40.12 & 39.40 & 43.65 \\
Hing-mBERT  & 28.52 & 26.64 & 23.81 & 27.53 & 67.27 & 100.00 & 78.96 & 49.65 & 43.54 & 29.95 & 53.63 & 29.71 \\
Hing-mBERT-Mixed  & 26.19 & 52.01 & 44.67 & 58.04 & 48.19 & 89.67 & 60.43 & 60.55 & 58.42 & 39.12 & 59.70 & 52.98 \\
XLM-R   & 56.70 & 99.97 & 95.58 & 89.67 & 77.30 & 100.00 & 50.77 & 47.00 & 54.93 & 56.73 & 42.53 & 40.37 \\
Hing-RoBERTa  & 20.17 & 37.22 & 51.56 & 99.68 & 98.87 & 100.00 & 46.33 & 49.93 & 19.92 & 13.96 & 13.82 & 15.34 \\
Hing-RoBERTa-Mixed  & 20.50 & 58.64 & 58.40 & 92.62 & 88.97 & 100.00 & 52.92 & 66.48 & 50.98 & 54.10 & 38.58 & 79.39 \\
\bottomrule
\end{tabular}}
\\
\caption{Layer-wise cross-lingual alignment accuracy (\%) across transformer layers (L1–L12) for multilingual and code-mixed adapted models using dot-product similarity with FAISS-based negative sampling between English (EN), Hindi (HI), and code-mixed (CM).}
\label{tab:dot_faiss_layerwise}
\end{table*}

\end{document}